\documentclass{article}

\usepackage[final]{corl_2022} 

\usepackage{microtype}
\usepackage{graphicx}
\usepackage{subfigure}
\usepackage{booktabs} 
\usepackage{comment}
\usepackage{hyperref}

\usepackage{amsmath}
\usepackage{amssymb}
\usepackage{mathtools}
\usepackage{amsthm}

\usepackage[capitalize,noabbrev]{cleveref}
\usepackage{mathtools}
\usepackage{algorithm}
\usepackage{algpseudocode}
\usepackage{float}
\usepackage{color}
\usepackage{colortbl}
\usepackage{rotating}
\usepackage{times}
\usepackage{epsfig}
\usepackage{enumitem}
\usepackage{makecell}
\usepackage{lineno}
\usepackage{tabularx}
\usepackage{xcolor}
\usepackage{multirow}
\usepackage{graphicx}
\usepackage{hhline}
\usepackage{textcomp,booktabs}
\usepackage{caption}
\usepackage{stmaryrd}
\usepackage[mathscr]{eucal}
\usepackage{lineno}
\usepackage[export]{adjustbox}

\def\0{{\bf 0}}
\def\1{{\bf 1}}

\theoremstyle{plain}

\theoremstyle{definition}

\theoremstyle{remark}

\usepackage[textsize=tiny]{todonotes}

\definecolor{red}{rgb}{0.95,0.4,0.4}
\definecolor{purered}{rgb}{1,0,0}
\definecolor{blue}{rgb}{0.4,0.4,0.95}
\definecolor{darkblue}{rgb}{0,0,0.8}
\definecolor{darkred}{rgb}{1,0,0}
\definecolor{darkgreen}{rgb}{0,0.5,0}
\definecolor{grey}{rgb}{0.6,0.6,0.6}
\definecolor{col1}{RGB}{232, 161, 148}
\definecolor{col2}{RGB}{148, 187, 232}
\definecolor{lightgrey}{rgb}{0.85,0.85,0.85}
\definecolor{lightlightgrey}{rgb}{0.9,0.9,0.9}
\definecolor{verylightBG}{rgb}{0.9,0.99,0.99}
\definecolor{darkgreen}{rgb}{0.3, 0.75, 0.3}

\usepackage{lipsum}

\title{Towards Long-Tailed 3D Detection}

\author{
  Neehar Peri\textsuperscript{1}, 
  Achal Dave\textsuperscript{1}, 
  Deva Ramanan\textsuperscript{1,2}$^*$,
  Shu Kong\textsuperscript{3}$^*$\\
  \textsuperscript{1}Robotics Institute, Carnegie Mellon University\\
  \textsuperscript{2}Argo AI, 
  \textsuperscript{3}Dept. of Computer Science and Engineering, Texas A\&M University\\
  \texttt{\{nperi, adave, deva\}@andrew.cmu.edu} \ \ \  {\tt shu@tamu.edu}
}

\begin{document}
\maketitle
\newcommand{\problem}{LT3D}
\newcommand{\eg}{e.g.}
\vspace{-5mm}

\begin{abstract}
Contemporary autonomous vehicle (AV) benchmarks have advanced techniques for training 3D detectors, particularly on large-scale LiDAR data. Surprisingly, although semantic class labels naturally follow a long-tailed distribution, these benchmarks focus on only a few {\tt common} classes (e.g., {\tt pedestrian} and {\tt car})
and neglect many {\tt rare} classes in-the-tail (e.g.,  {\tt debris} and {\tt stroller}).
However, in the real open world, AVs must still detect {\tt rare} classes to ensure safe operation. 
Moreover, semantic classes are often organized within a hierarchy, e.g., tail classes such as {\tt child} and {\tt construction-worker} are arguably subclasses of {\tt pedestrian}. 
However, such hierarchical relationships are often ignored, which may yield misleading estimates of performance and missed opportunities for algorithmic innovation.
We address these challenges by formally studying the problem of \emph{Long-Tailed 3D Detection} (\problem), which evaluates on {\em all} classes, including those in-the-tail. 
We evaluate and innovate upon popular 3D detectors, such as CenterPoint and PointPillars, adapting them for \problem.
We develop hierarchical losses that promote feature sharing across common-vs-rare classes, as well as improved detection metrics that award partial credit to ``reasonable" mistakes respecting the hierarchy (e.g., mistaking a {\tt child} for an {\tt adult}). Finally, we point out that fine-grained tail class accuracy is particularly improved via  {\em multimodal fusion} of RGB images with LiDAR; simply put, fine-grained classes are challenging to identify from sparse (LiDAR) geometry alone, suggesting that multimodal cues are crucial to  long-tailed 3D detection. Our modifications improve accuracy by 5\% AP on average for all classes, and dramatically improve AP for {\tt rare} classes (e.g., {\tt stroller} AP improves from 0.1 to 31.6)! Our code is available on \href{https://github.com/neeharperi/LT3D}{GitHub}.


\end{abstract}



\section{Introduction}
\label{sec:intro}
3D object detection is a key component in many robotics systems such as autonomous vehicles (AVs)~\cite{geiger2012we, caesar2020nuscenes}. 
To facilitate research in this space, the AV industry has released large-scale 3D annotated multimodal datasets~\cite{caesar2020nuscenes, chang2019argoverse, sun2020waymo}.
However, these datasets benchmark on only a few {\tt common} classes such as {\tt pedestrian} and {\tt car}.
In the real open world, safe navigation~\cite{taeihagh2019governing, wong2020identifying} requires AVs to reliably detect {\tt rare}-class objects such as {\tt child} and {\tt stroller}.
This motivates {\em Long-Tailed 3D Detection} (\problem), a problem requiring detecting objects from both {\tt common} and {\tt rare}  classes. 

\begin{figure*}[t]
\centering
\small
\hspace{-74mm} {\tt Few} \hspace{10mm} {\tt Medium} \hspace{10mm} {\tt Many}\\
\includegraphics[width=.46\linewidth, clip, trim={0cm 0cm 0cm 0cm}]{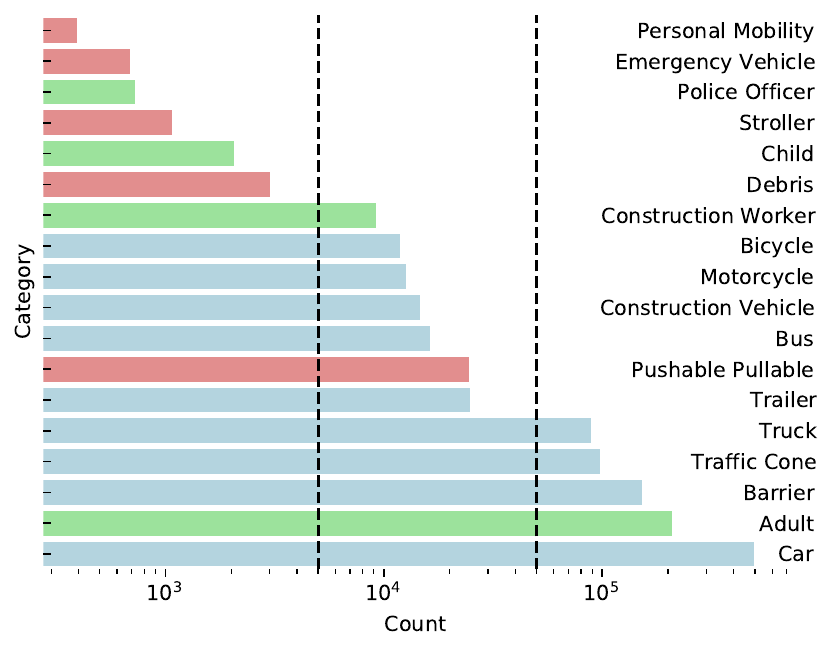} \hfill
\includegraphics[width=.47\linewidth]{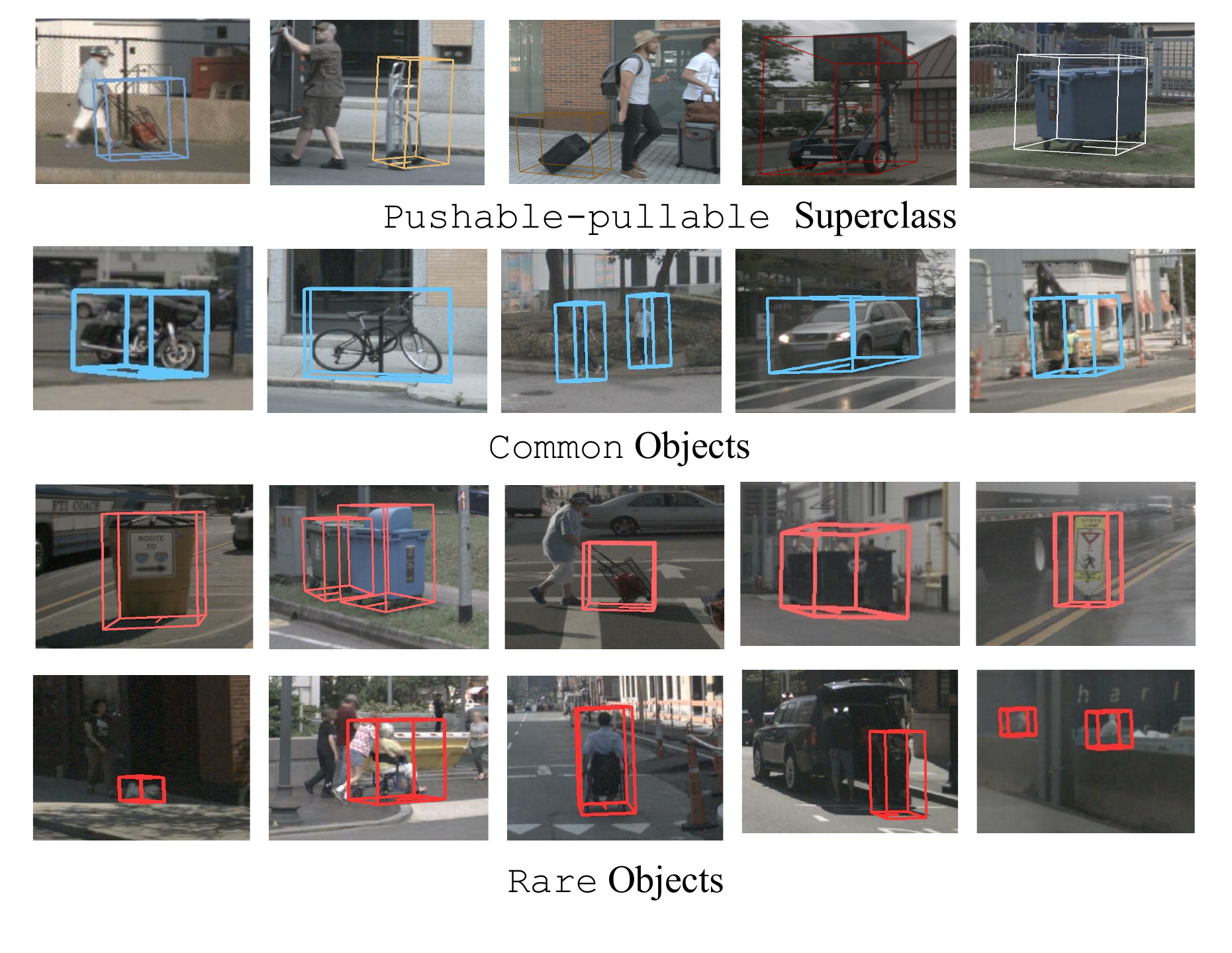} \\
\caption{\small
According to the histogram of per-class object counts (on the {\bf left}), 
the nuScenes benchmark focuses on the common classes in {\color{cyan} cyan} (e.g., {\tt car} and {\tt barrier}) but ignores rare ones in {\color{red} red} (e.g., {\tt stroller} and {\tt debris}). 
In fact, the benchmark creates a superclass {\tt pedestrian} by grouping multiple classes in {\color{darkgreen} green}, including the common class {\tt adult} and several rare classes (e.g., {\tt child} and {\tt police-officer}); this complicates the analysis of detection performance as {\tt pedestrian} performance is dominated by {\tt adult}.
Moreover, the ignored superclass {\tt pushable-pullable} also contains diverse objects such as {\tt shopping-cart}, {\tt dolly}, {\tt luggage} and {\tt trash-can} as shown in the top row (on the {\bf right}). We argue that AVs should also detect {\tt rare} classes as they can affect AV behaviors.
Following \cite{openlongtailrecognition}, we report performance for three groups of classes based on their cardinality (split by dotted lines): {\tt Many}, {\tt Medium}, and {\tt Few}.}
\label{fig:histogram}
\vspace{-5mm}
\end{figure*}

{\bf Status Quo}.
Among contemporary AV datasets,
nuScenes~\cite{caesar2020nuscenes} has exhaustively annotated objects of various classes crucial to AVs (Fig.~\ref{fig:histogram}) and organized them with a semantic hierarchy (Fig.~\ref{fig:hierarchy}).
As it focuses on only a few ({\tt common}) classes, prior works miss opportunities to exploit this semantic hierarchy during training. We argue that these benchmarking protocols are flawed because detecting fine-grained classes is useful for downstream tasks such as motion planning. This motivates us to study \problem\ by re-purposing {\em all} annotated classes in nuScenes.

\textbf{Protocol}. 
\problem \ requires 3D localization and recognition of objects from each of the {\tt common} (e.g., {\tt adult} and {\tt car}) and {\tt rare} classes (e.g, {\tt child} and {\tt stroller}).
Moreover, for safety-critical robots such as autonomous vehicles, we believe detecting but mis-classifying {\tt rare} objects (e.g., mis-classifying a {\tt child} as an {\tt adult}) is preferable to failing to detect them at all. 
Therefore, we propose a new metric to quantify the severity of classification mistakes in \problem \ that exploits inter-class relationships to award partial credit (Fig.~\ref{fig:hierarchy}).
We use both the standard and proposed metrics to evaluate 3D detectors on all classes.

{\bf Technical Insights}.
To address \problem, we first retrain state-of-the-art LiDAR-based 3D detectors on
{\em all} classes.
Naively retraining detectors produces poor performance on {\tt rare} classes (e.g., yielding 0.1 AP on {\tt child} and 0.1 AP on {\tt stroller}).
We propose several algorithmic innovations to improve these results. 
First, to encourage feature sharing across common-vs-rare classes, we learn a single feature trunk, adding in hierarchical coarse classes that ensure features will be useful for both {\tt common} and {\tt rare} classes. 
Second, we find that LiDAR data is simply too impoverished for even humans to recognize certain tail objects that tend to be small, such as {\tt strollers}. We explore multimodal fusion, and introduce a simple approach that post-processes single-modal 3D detections from LiDAR and RGB inputs, filtering away detections that are inconsistent across modalities. Our innovations significantly improve performance on \problem \ by 5 \% AP on average, 
greatly boosting performance when allowing for partial credit  (e.g., achieving 16.9 / 38.8 AP for {\tt child} / {\tt stroller}).

{\bf Contributions}.
We make three major contributions.
First, we formulate the problem \problem, emphasizing detection of both {\tt common} and {\tt rare} classes in safety-critical AVs.
Second, we design \problem's benchmarking protocol and develop a supplemental metric that awards partial credit depending on the severity of misclassifications (e.g., misclassifying {\tt child}-vs-{\tt adult} is less problematic than misclassifying {\tt child}-vs-{\tt car}). 
Third, we propose several architecture-agnostic approaches to \problem, including a simple multimodal fusion technique that uses RGB-based detections to filter out false-positive LiDAR-based detections, leading to significant improvement of {\tt rare}-class detection.

\section{Related Works}
{\bf 3D Object Detection}. Contemporary benchmarks, often in the AV setting, favor LiDAR-based detectors, emphasizing {\tt common} classes and ignoring {\tt rare} ones.
Approaches to 3D detection usually adopt an anchor-based model architecture that defines per-class shapes to guide class-aware object detection~\cite{lang2019pointpillars, zhu2019class, hu2019wysiwyg, yan2018second, wang2019exploit}. 
A recent \textit{anchor-free} model, CenterPoint~\cite{yin2020center} achieves the state-of-the-art for LiDAR-based 3D object detection. Specifically, it learns to predict an object's center and estimates the 3D shape for each detected object's center.
Existing LiDAR-based 3D detectors exclusively focus on data from {\tt common} classes \cite{lang2019pointpillars, zhu2019class, yin2020center} and do not study how to detect {\tt rare} classes. 
RGB-based 3D detectors underperform LiDAR-based methods because the monocular RGB input does not provide reliable 3D measures (unlike LiDAR). As a result, RGB-based 3D detectors are not widely adopted.
However, in exploring \problem\, we find that RGB-detectors shine for detecting objects of {\tt rare} classes. Importantly, multimodal fusion significantly improves \problem.

\textbf{Multimodal 3D Detection}. 
Conventional wisdom suggests that fusing multimodal cues, particularly using LiDAR and RGB, can improve 3D detection. Intuitively, LiDAR faithfully measures the 3D world (although it has notoriously sparse point returns), and RGB is  high-resolution (but lacks 3D information). Multimodal fusion for 3D detection is an active field of exploration. Existing methods suggest different ways to fuse the two modalities. Proposed methods encode separate modalities and fuse object proposals~\cite{chen2017multiview,ku2018joint, yoo203dcvf, bai22transfusion, chen2021multimodal, gupta2023far3det}, augment LiDAR points with either RGB features~\cite{sindagi2019mvxnet}, augment RGB images with LiDAR points~\cite{you2020pseudoLiDAR} or add semantic information obtained by processing RGB inputs~\cite{vora2020pointpainting, yin2021multimodal}. Others
propose stage-wise methods that first detect boxes over images and localize in 3D with LiDAR~\cite{qi2018frustum} and fuse detections from single-modal detectors~\cite{xu2018pointfusion, pang2009clocs}.
While the above methods have not been tested for \problem, our work shows that RGB is a key modality for \problem.

\textbf{Long-Tailed Perception} (LTP). 
Real-world data tends to follow long-tailed class distributions~\cite{reed2001pareto}, i.e., a few classes are dominant in the data, while many others are rarely seen. LTP is a long-standing problem in the literature~\cite{openlongtailrecognition}. 
It has been widely studied through the lens of image classification and requires training on class-imbalanced data, aiming for high accuracy averaged across imbalanced classes~\cite{openlongtailrecognition, zhang2021deep, alshammari2022long}.
Existing methods propose reweighting losses~\cite{cui2019class, khan2017cost, cao2019learning, khan2019striking, huang2019deep, zhang2021distribution}, rebalancing data sampling~\cite{drummond2003c4, chawla2002smote, han2005borderline}, balancing gradients computed from imbalanced classes~\cite{tang2020long}, and balancing network weights~\cite{alshammari2022long}.
Others study LTP through the lens of 2D object detection over RGB images~\cite{gupta2019lvis}. Compared to 2D image-based recognition, 3D long-tailed detection has unique opportunities and challenges because sensors such as LiDAR directly provide geometric and ego-motion cues that are difficult to extract from 2D images.
2D detectors must detect objects of different scales due to perspective image projection, dramatically increasing the complexity of the output space (e.g., requiring more anchor boxes). 
In contrast, 3D objects do not exhibit as much scale variation, but far-away objects tend to be sparse, imposing different challenges. Finally, 3D detectors often use class-aware heads (i.e. each class has its own binary classifier) while 2D long-tail recognition typically use shared softmax heads (which may make it easier to enforce hierarchies, as explained above). 
To the best of our knowledge, long-tailed 3D detection (\problem) has not yet been explored.
In \problem, we find a unique challenge: rare classes are not only infrequent but are also difficult to distinguish using LiDAR alone.
This motivates us to use RGB to complement LiDAR. We find using both RGB (for better recognition) and LiDAR (for better 3D localization) helps detect {\tt rare} classes.

\section{\problem: Methodology}
\label{sec:methods}

To approach \problem, we first retrain SOTA 3D detectors on {\em all} classes, including LiDAR-based detectors (PointPillars~\cite{lang2019pointpillars}nand CenterPoint~\cite{yin2020center}), an RGB-based detector (FCOS3D~\cite{wang2021fcos3d}), and multimodal detector (TransFusion~\cite{bai22transfusion}).
We further introduce several modifications that consistently improve their \problem \ performance. Please refer to the supplement for training details.

\textbf{Grouping-Free Detector Head}. Extending existing 3D detectors to train with more classes is surprisingly challenging. 
Many contemporary networks use a multi-head architecture that groups classes of similar size and shape to facilitate efficient feature sharing. For example, CenterPoint groups {\tt pedestrian} and {\tt traffic-cone} since these objects are both tall and skinny.
However, multi-headed grouping strategies may not work for diverse classes like {\tt pushable-pullable} and {\tt debris} and are difficult to scale for a large number of classes.  Therefore, we first consider making each class its own group to avoid hand-crafted grouping heuristics. 
However, the multi-head architecture has per-class detectors that consist of multiple layers with lots of parameters, hence learning them easily overfits to rare-classes. Our final solution is to merge all classes into a single group with a proportionally heavier detector head to simplify training. Our group-free (i.e. single-head) architecture has a shared backbone across all classes, and each class has only one linear layer as its detector. This significantly reduces the number of parameters  and allows learning the shared feature backbone collaboratively with all classes, effectively mitigating overfitting to rare-classes. Adding a new class is as simple as adding a single linear layer to the detector head. Our grouping-free detector head achieves improved accuracy over grouping-based methods, as shown in the supplement. 


\textbf{Training with Semantic Hierarchies}. 
nuScenes defines a semantic hierarchy (Fig.~\ref{fig:hierarchy}) for all classes, grouping semantically similar classes under coarse-grained categories. We leverage this hierarchy during training.
Specifically, we train detectors to predict three labels for each object: its fine-grained label (e.g., {\tt child}, its coarse class (e.g., {\tt pedestrian}), and the root class {\tt object}.
We adopt a grouping-free detector head that outputs separate ``multitask'' heatmaps for each class, and use a per-class sigmoid focal loss rather than multi-class cross-entropy loss. It is worth noting that this simple ``multitask'' learning strategy does not necessarily enforce a hierarchy, hence can extend to more complex label relationships.
Crucially, because we do not employ softmax losses, adding a {\tt vehicle} heatmap does not directly interfere with the {\tt car} heatmap (as they would with a multi-class softmax loss). However, this might produce repeated detections on the same test object. We address that by simply ignoring coarse detections at test time. We explore alternatives in the supplement and conclude that they achieve similar \problem \ performance.
Perhaps surprisingly, this training method improves detection performance not only for {\tt rare} classes, but also for {\tt common} classes. 

\textbf{Augmentation Schedule}. Class-balanced resampling is a common technique in learning with long-tailed classes. This augmentation strategy increases the number of {\tt rare} objects seen in training but skews the class distribution and leads to more false positives for {\tt rare} classes in inference. Prior works~\cite{vora2020pointpainting, bai22transfusion} suggest disabling class-balanced resampling for the last few training epochs to better match the real class distribution, reducing false positives. We validate this approach in training 3D detectors and find that it often improves performance for {\tt rare} classes at the cost of {\tt common} classes. 

\begin{figure}[t]
\centering
\small
\hspace{7mm} LiDAR-based Detections \hspace{13mm} RGB-based Detections \hspace{10mm} Filtered LiDAR-based Detections \\
\includegraphics[width=\linewidth, height=3.1cm]{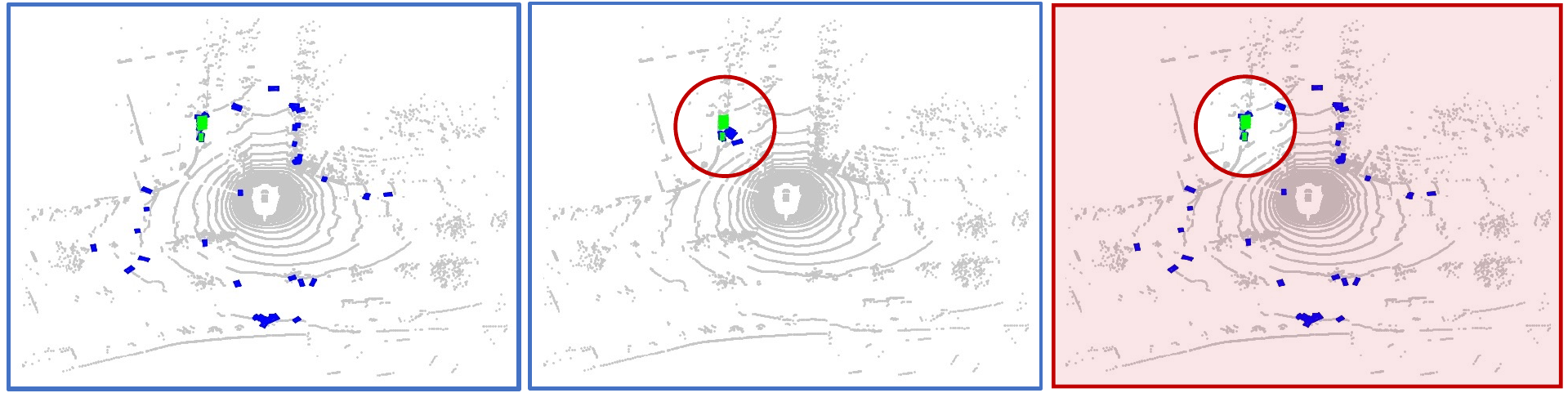}
\caption{\small
Multimodal filtering effectively removes high-scoring false-positive LiDAR detections.
The green boxes are ground-truth {\tt strollers}, while the blue boxes are {\tt stroller} detections from our best performing models, liDAR-based detector CenterPoint~\cite{yin2020center} ({\bf left}) and RGB-based detector FCOS3D~\cite{wang2021fcos3d} ({\bf mid}). The final filtered result removes LiDAR detections not within $m$ meters of any RGB detection ({\bf right}).} 
\label{fig:filtering}
\vspace{-2mm}
\end{figure}

\textbf{Multimodal Fusion by Filtering}. 
Small fine-grained classes are challenging to identify from sparse (LiDAR) geometry alone, suggesting that multimodal cues can improve long-tailed detection. We evaluate several multimodal fusion algorithms, but find a simple strategy of post-hoc filtering to work remarkably well.
Although LiDAR-based detectors are widely adopted for 3D detection, we find that they produce many high-scoring false positives (FPs) for rare classes due to misclassification. We focus on  removing such FPs. 
To this end, we use an RGB-based detector to filter out high-scoring false-positive LiDAR detections by leveraging two insights: (1) LiDAR-based 3D-detectors are accurate w.r.t 3D localization and yield high recall (though classification is poor), and (2) RGB-based 3D-detections are accurate w.r.t recognition (though 3D localization is poor). 
Fig.~\ref{fig:filtering} demonstrates this filtering strategy. For each RGB-based detection, we keep LiDAR-based detections within a radius of $m$ meters and remove all the others (that are not close to any RGB-based detections).
We use FCOS3D~\cite{wang2021fcos3d} as the RGB-based detector in this work.

\section{\problem: Evaluation Protocol}
\label{sec:protocol}

Conceptually, \problem \ extends the traditional 3D detection problem, which focuses on identifying objects from  $K$ {\tt common} classes, by further requiring detection of $N$ {\tt rare} classes.
As \problem\ emphasizes detection performance on {\em all} classes,
we report the metrics for three groups of classes based on their cardinality (Fig.~\ref{fig:histogram}-left): {\em many} ($>$50k instances per class), {\em medium} (5k$\sim$50k), and {\em few} ($<$5k).
We describe the metrics below.

\textbf{Standard Detection Metrics}.
Mean average precision (mAP) is an established metric for object detection~\cite{everingham2015pascal, geiger2012we, lin2014coco}.
For 3D detection on LiDAR sweeps, a true positive (TP) is defined as a detection that has a center distance within a distance threshold on the ground-plane to a ground-truth annotation~\cite{caesar2020nuscenes}.
mAP computes the mean of AP over classes, where per-class AP is the area under the precision-recall curve, and distance thresholds of [0.5, 1, 2, 4] meters.

\textbf{Hierarchical Mean Average Precision (mAP$_{H}$)}.
For safety critical applications, although correctly localizing and classifying an object is ideal, detecting but misclassifying {\em some} object is more desirable than a missed detection (e.g., detect but misclassify a {\tt child} as {\tt adult} versus not detecting this {\tt child}). 
Therefore, we introduce hierarchical AP (AP$_H$) which  considers such semantic relationships across classes to award partial credit. 
To encode these relationships between classes, we leverage the semantic hierarchy (Fig.~\ref{fig:hierarchy}) defined by nuScenes. We derive partial credit as a function of semantic similarity using the least common ancestor (LCA) distance metric. Hierarchical metrics have been proposed for image classification \cite{russakovsky2015imagenet}, but have not been explored for object detection. 
Extending this metric for object detection is challenging because we must consider how to jointly evaluate semantic and spatial overlap. For clarity, we will describe the procedure in context of computing AP$_H$ for some arbitrary class $C$. 

{\bf LCA=0}: Consider the predictions and ground-truth boxes for $C$. Label the set of predictions that overlap with ground-truth boxes for $C$ as true positives. Other predictions are false positives. {\em This is identical to the standard AP metric.}

{\bf LCA=1}: Consider the predictions for $C$, and ground-truth boxes for $C$ and all sibling classes of $C$ (that have LCA distance to $C$ of 1). Label the set of predictions that overlap a ground-truth box of $C$ as a true positive. Label the set of predictions that overlap sibling classes as {\tt ignored}~\cite{lin2014coco}. 
All other predictions for $C$ are false positives.

{\bf LCA=2}: Consider the predictions for $C$ and ground-truth boxes for $C$ and all sibling classes of $C$ (that have LCA distance to $C$ less than $2$. For nuScenes, this includes all classes.) Label the set of predictions that overlap ground-truth boxes for $C$ as true positives. Label the set of predictions that overlap other classes as {\tt ignored}. All other predictions for $C$ are false positives.

\section{Experiments}
We conduct experiments to better understand the \problem \ problem, and gain insights by validating our techniques described in Section~\ref{sec:methods}.
Specifically, we aim to answer the following questions:\footnote{Answers: yes, yes, yes, yes.}
\begin{enumerate}[noitemsep,  topsep=0pt] 
\item Are {\tt rare} classes more difficult to detect than {\tt common} classes?
\item Are objects from {\tt rare} classes sufficiently localized but mis-classified?
\item Does training with the semantic hierarchy improve detection performance for \problem?
\item Does multimodal fusion help detect  {\tt rare} classes?
\end{enumerate}

We start this section by introducing the model architecture, implementation and dataset.

{\bf Model Architecture}.
For LiDAR-based 3D detectors, we use PointPillars~\cite{lang2019pointpillars} and CenterPoint~\cite{yin2020center}, which are widely benchmarked in the literature.
For fusion-based 3D detectors, we use TransFusion~\cite{bai22transfusion}, a recently released state-of-the-art method. TransFusion proposes an end-to-end DETR-like approach~\cite{detr2020carion} for multimodal 3D detection.


{\bf Implementation}.
We use the PyTorch toolbox~\cite{paszke2019pytorch} to train all models for 20 epochs with the Adam optimizer \cite{kingma2015adam} and a one-cycle learning rate scheduler \cite{smith2017cyclical}. 
In training, we adopt data augmentation techniques introduced by~\cite{yin2020center}. When using the introduced data augmentation schedule (cf. Section~\ref{sec:methods}), we train models for 15 epochs with data augmentation enabled, and 5 epochs without. We further describe our implementation in the supplement.

\begin{figure}[t]
\centering
\includegraphics[width=0.54\linewidth, clip, trim={0cm 0cm 0cm 0cm}]{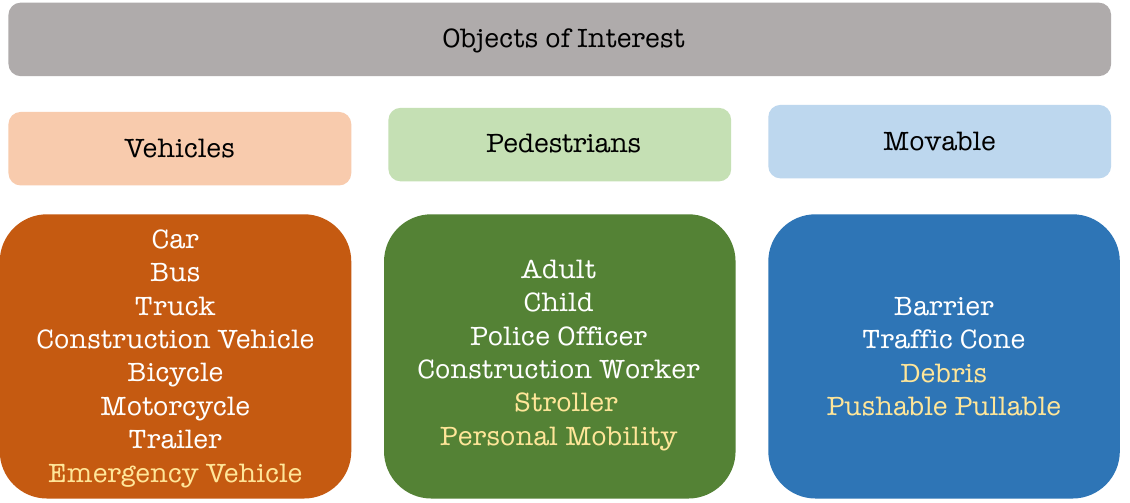} \hfill
\includegraphics[width=0.4\linewidth]{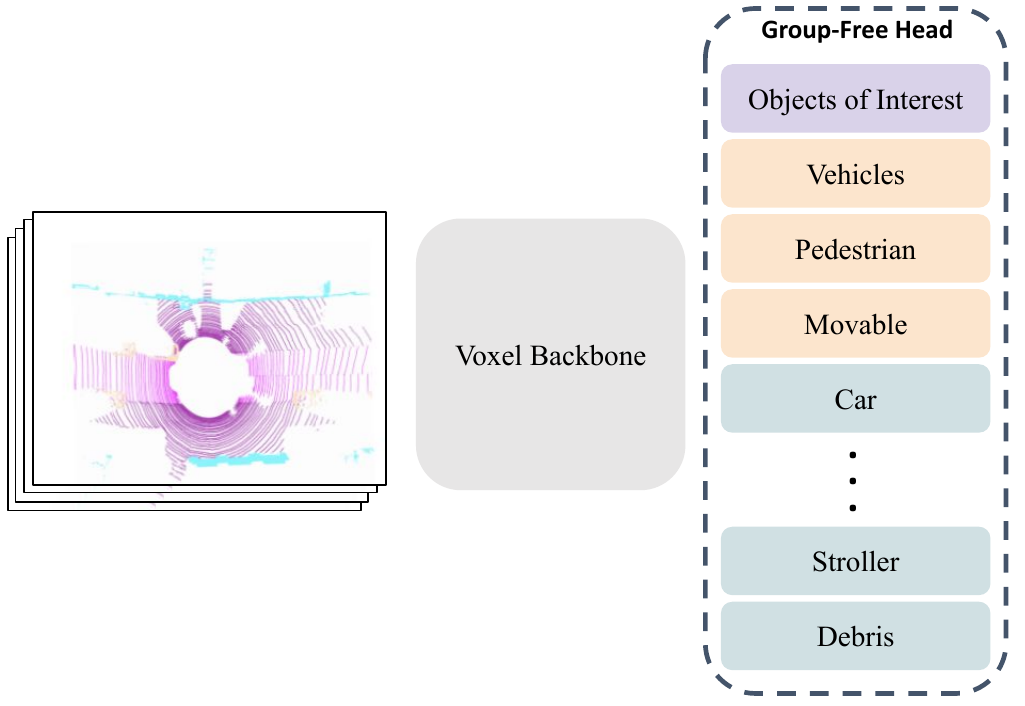} 
\caption{\small
nuScenes defines a semantic hierarchy (on the {\bf left}) for all annotated classes (Fig.~\ref{fig:histogram}). We highlight {\tt common} classes in white and {\tt rare} classes in gold. The standard nuScenes benchmark makes two choices for dealing with rare classes: (1) ignore them (e.g., {\tt stroller} and {\tt pushable-pullable}), or (2) group them into coarse-grained classes (e.g., {\tt adult}, {\tt child}, {\tt construction-worker}, {\tt police-officer} are grouped as {\tt pedestrian}). Since the {\tt pedestrian} class is dominated by {\tt adult} (Fig.~\ref{fig:histogram}), the standard benchmarking protocol masks the challenge of detecting rare classes like {\tt child} and {\tt police-officer}. We leverage this hierarchy  during training (on the {\bf right}) by predicting class labels at {\em multiple} levels of the hierarchy. Specifically, we train detectors to predict three labels for each object: its fine-grained label (e.g., {\tt child}, its coarse class (e.g., {\tt pedestrian}), and the root-level class {\tt object}. This means that the final vocabulary of classes is no longer mutually exclusive, complicating the application of multi-class softmax losses. To address this, use a sigmoid focal loss that learns separate spatial heatmaps for each class.
\vspace{-2mm}
}
\label{fig:hierarchy}
\end{figure}

{\bf Datasets.}
We use nuScenes~\cite{caesar2020nuscenes} and Argoverse 2.0 (AV2)~\cite{wilson2021argoverse} to explore \problem.
Both have fine-grained classes (18 and 26 classes in nuScenes and AV2 respectively) that follow long-tailed distributions. To quantify the long-tail, we use the imbalance factor (IF) defined as the ratio between the numbers of annotations of the max-class and min-class~\cite{cao2019learning}; nuScenes and AV2 have IF=1670 and 2500 respectively -- significantly more imbalanced than  existing long-tail image recognition benchmarks, e.g., iNaturalist (IF=500)~\cite{van2018inaturalist} and ImageNet-LT (IF=1000)~\cite{liu2019large}.
NuScenes arranges classes in a semantic hierarchy (Fig.~\ref{fig:hierarchy}); AV2 does not provide a semantic hierarchy but we construct one based on the nuScenes' hierarchy.
Following prior work, we use their official train-sets for training and evaluate on the their official val-sets. We focus on nuScenes in the main paper and AV2 in the supplement. Our primary conclusions hold for both datasets.


{
\setlength{\tabcolsep}{1.23em} 
\begin{table}[t]
\small
\centering
\caption{\small 
{\bf Benchmarking detectors for \problem} (measured by mAP). We present several salient conclusions. 
First, training with the semantic hierarchy improves all methods for \problem, e.g., improving  by 1\% AP averaged over {\tt All} classes.
Second, multimodal filtering yields between 4$\sim$11 AP improvement on {\tt Medium} and {\tt Few} classes! This is surprising given that FCOS3D is a less powerful 3D detector on its own. Importantly, we do not modify FCOS3D for \problem. Interestingly, it also improves multimodal detectors (1.6\% AP improvement for TransFusion on {\tt All} classes), demonstrating the importance of using RGB to improve \problem\ with better recognition.
Third, perhaps surprisingly, post-hoc multimodal filtering of LiDAR-only detector CenterPoint with RGB-only detector FCOS3D performs the best, surpassing the multi-modal TransFusion model.
Lastly, data augmentation schedules do not necessarily improve \problem \ performance, demonstrating the challenge of 3D detection in the long-tail.
 }
\begin{tabular}{l c c c c c c c c c c c}
\toprule
\multirow{1}{*}{Method} & \multirow{1}{*}{Multimodal} & \multicolumn{1}{c}{\tt Many} & \multicolumn{1}{c}{\texttt{Medium}} & \multicolumn{1}{c}{\texttt{Few}} & \multicolumn{1}{c}{\tt All} \\ 
\midrule
\multirow{1}{*}{FCOS3D (RGB-only) \cite{wang2021fcos3d}}        &           &  39.0            & 23.3         &   2.9        &   20.9       \\
\Xhline{3\arrayrulewidth}
\multirow{1}{*}{PointPillars (LiDAR-only) \cite{lang2019pointpillars}}                      &             &  64.2            & 28.4           &   3.4        &   30.0     \\ 
\multirow{1}{*}{$\indent$ + Hierarchy}                                         &             &  \textbf{66.4}   & 30.4           &   2.9        &   31.2        \\
\multirow{1}{*}{\quad\quad $\indent$ w/ Data Aug.}                                        &             &  54.4            & 24.2           &   1.8        &   25.1       \\ 
\multirow{1}{*}{\quad\quad $\indent$ w/ Multimodal Filtering}                                        & \checkmark  &  66.2            & \textbf{41.0}  &   \textbf{4.4}        &   \textbf{35.8}      \\


\midrule
\multirow{1}{*}{CenterPoint (LiDAR-only) \cite{yin2020center}}                       &            &  76.4             & 43.1         &   3.5        &   39.2        \\ 
\multirow{1}{*}{$\indent$ + Hierarchy}          &                                    &  \textbf{77.1}             & 45.1         &   4.3         &   40.4      \\
\multirow{1}{*}{\quad\quad $\indent$ w/ Data Aug.}         &                                    &  73.8             & 44.5         &   7.4       &   40.3      \\ 
\multirow{1}{*}{\quad\quad $\indent$ w/ Multimodal Filtering}         &   \checkmark                       &  \textbf{77.1}             & \textbf{49.0}          &   \textbf{9.4}      &   \textbf{43.6}       \\ 

\midrule
\multirow{1}{*}{TransFusion-L (LiDAR-only)\cite{bai22transfusion}}                   &                  &  68.5            & \textbf{42.8}          &   8.4        &   38.5       \\ 
\multirow{1}{*}{\quad\quad $\indent$ w/ Multimodal Filtering}                                &  \checkmark      &  73.2            & 42.5          &   8.3        &  39.6      \\ 
\multirow{1}{*}{TransFusion (LiDAR + RGB)}                                    &  \checkmark      &  \textbf{73.9}            & 41.2          &   \textbf{9.8}        &   39.8       \\
\multirow{1}{*}{\quad\quad $\indent$ w/ Data Aug.}                                &  \checkmark      &  73.4             & 40.9         &   8.2       &   39.0      \\ 
\multirow{1}{*}{\quad\quad $\indent$ w/ Multimodal Filtering}                                &  \checkmark      &  \textbf{73.9}            & 42.5          &   9.1        &  \textbf{40.1}      \\ 
\bottomrule
\end{tabular}
\label{tab:lt3d}

\end{table}
}


{\bf Retraining state-of-the-art 3D detectors for \problem}. We  retrain several 3D detectors, namely FCOS3D~\cite{wang2021fcos3d}, PointPillars~\cite{lang2019pointpillars} and CenterPoint~\cite{yin2020center}. FCOS3D operates on monocular images. The other three detectors take an aggregated stack of ten LiDAR-sweeps as input. All four models predict 3D bounding boxes for 18 classes as defined by the nuScenes \problem \ protocol. As shown in Table \ref{tab:lt3d}, LiDAR-based detectors that perform well on {\tt common} classes tend to also perform well on {\tt rare} classes. 

{\bf Training with Semantic Hierarchy.} Next, we modify our LiDAR-based detectors to jointly predict class labels at different levels of the semantic hierarchy. For example, we modify the detector to additionally classify {\tt stroller} as {\tt pedestrian} and {\tt object}. The semantic hierarchy naturally groups classes based on shared attributes and may have complementary features. Moreover, training with the semantic hierarchy allows {\tt rare} classes within each group to learn better features by sharing with {\tt common} classes. This approach is generally effective, as shown in Table \ref{tab:lt3d}, improving accuracy for classes with {\tt Many} examples by 2\%, {\tt Medium} examples by 2\% and {\tt Few} examples by 1\% AP. 

{\bf Data Augmentation Schedule}. Prior works \cite{xu2018pointfusion, bai22transfusion} suggest disabling copy-paste augmentation for the last few epochs of training to reduce the number of false positive detections. We validate this claim for various detector architectures and find that although it seems to help {\tt rare} classes by 3\% AP, but hurts {\tt common} classes by 4\% AP (c.f. CenterPoint). 

{\bf Multimodal Fusion via Filtering.} Detecting {\tt rare} classes from LiDAR-only is challenging since its difficult to recognize objects from sparse LiDAR points alone. As a result, LiDAR-detectors often have many high-scoring FPs (Fig.~\ref{fig:filtering}), resulting in low AP. 
Using RGB detections to filter the LiDAR detections results in significant performance improvement on {\tt rare} classes, improving by 4$\sim$11 AP on classes with {\tt Medium} and {\tt Few} examples for all models (Table~\ref{tab:lt3d})!

{\bf End-to-End Multimodal Methods}. Since multimodal cues significantly improve \problem, we are motivated to explore end-to-end approaches. Specifically, we retrain TransFusion~\cite{bai22transfusion} on all 18 classes. 
We retrain TransFusion, downsampling the RGB images by a factor of 2 to fit the model in GPU memory. Surprisingly, lidar-only and multi-modal variants of TransFusion perform worse than CenterPoint for \problem. Further hyperparameter tuning and full-resolution training may help.
Lastly, our multimodal filtering strategy still improves this end-to-end fusion methods slightly, e.g., it increases mAP on {\tt All} classes by 0.3\% AP for TransFusion (cf. Table~\ref{tab:lt3d}). We also find that applying multimodal filtering on the LiDAR-only branch of TransFusion yields similar performance to the multimodal model, suggested that TransFusion is simply learning a multimodal filtering function.

{
\setlength{\tabcolsep}{0.34em} 
\begin{table}[t]
\small
\centering
\caption{\small 
{\bf Diagnosis using the mAP$_H$ metric on selected classes}.
We analyze the best-performing LiDAR-only model CenterPoint and multimodal model TransFusion, with / without our hierarchical loss (\emph{hier.}) and mutimodal filtering technique ({\em filtering}).
Comparing the rows of LCA=0 for with and without our techniques (for CenterPoint and TransFusion respectively), we see our techniques bring significantly improvements on classes with  {\color{blue}{\tt medium}} and {\color{blue}{\tt few}} examples such as {\tt construction-vehicle} (CV), {\tt bicycle}, {\tt motorcycle} (MC), {\tt construction-worker} (CW), {\tt stroller}, and {\tt pushable-pullable} (PP).
Moreover, performance increases significantly from LCA=0 to LCA=1 compared against LCA=1 to LCA=2, (Table \ref{tab:ap_h}), confirming that objects from {\tt rare} classes are often detected but misclassified as some sibling classes.  
}
\begin{tabular}{c l c c c c c c c c c c c c c}
\toprule
  \multirow{1}{*}{Method}  &\multirow{1}{*}{$mAP_H$}  & \multicolumn{1}{c}{Car} & \multicolumn{1}{c}{Adult} & \multicolumn{1}{c}{Truck} & \multicolumn{1}{c}{\color{blue}CV} &  \multicolumn{1}{c}{\color{blue}Bicycle} & \multicolumn{1}{c}{\color{blue}MC} &  \multicolumn{1}{c}{\color{blue}Child} & \multicolumn{1}{c}{\color{blue}CW} & \multicolumn{1}{c}{\color{blue}Stroller} & \multicolumn{1}{c}{\color{blue}PP} \\ 

\midrule
\multirow{3}{*}{CenterPoint (OTS)}    & \multirow{1}{*}{LCA=0}    &     82.4    &   81.2         &    49.4      &      19.7             &   33.6         &    48.9          &      0.1             &   14.2        &    0.1        &      21.7     \\
  & \multirow{1}{*}{LCA=1}                                  &     83.9    &   82.0         &    58.7      &      20.5             &   35.2         &    50.5          &      0.1             &   18.3        &    0.1        &      22.0     \\
    & \multirow{1}{*}{LCA=2}                                &     84.0    &   82.4         &    58.8      &      20.7             &   36.4         &    51.0          &      0.1             &   19.5        &    0.1        &      22.6     \\
\midrule
\multirow{3}{*}{CenterPoint (Group-Free)}    & \multirow{1}{*}{LCA=0}        &      88.1      &   86.3         &    62.7           &      24.5      &   48.5        &    62.8          &      0.1        &   22.2         &    4.3         &      32.7    \\
  & \multirow{1}{*}{LCA=1}                                                   &      89.0      &   87.1         &    71.6           &      26.7      &   50.2        &    64.7          &      0.1        &   29.4         &    4.5         &      32.9    \\
    & \multirow{1}{*}{LCA=2}                                                 &      89.1      &   87.5         &    71.7           &      26.8      &   51.1        &    65.2          &      0.1        &   30.5         &    4.8         &      33.4    \\
\midrule
    \multirow{2}{*}{CenterPoint (Group-Free)}  & \multirow{1}{*}{LCA=0}                    &      {\bf 88.6}     &  {\bf 86.9}      &   {\bf 63.4}           &      25.7          &     50.2       &    63.2       &      0.1           &   25.3         &    8.7     &    36.8    \\ 
    \multirow{2}{*}{\em w/ Hierarchy}   & \multirow{1}{*}{LCA=1}              &      {\bf 89.5}     &   {\bf 87.6}      &    {\bf 72.4}           &      27.5          &     52.2       &    65.2       &      0.1           &   32.4         &    9.4     &    37.0    \\
     & \multirow{1}{*}{LCA=2}                                                 &      {\bf 89.6}     &   {\bf 88.0}      &    {\bf 72.5}           &      27.7          &     53.2       &    65.7       &      0.1           &   34.0         &    9.8     &    37.6    \\
\midrule
   \multirow{2}{*}{CenterPoint (Group-Free)}  & \multirow{1}{*}{LCA=0}                     &      88.5             &   86.6         &    {\bf 63.4}           &      {\bf 29.0}             &   {\bf 58.5}         &    {\bf 68.2}          &      {\bf 5.3}             &   {\bf 35.8}         &    {\bf 31.6}          & {\bf 39.3}             \\ 
 \multirow{2}{*}{\em w/ Hier. \& MM. Filtering} & \multirow{1}{*}{LCA=1}                     &      89.4             &   87.4         &    {\bf 72.4}           &     {\bf 31.3}             &   {\bf 61.2}         &    {\bf 69.7}          &      15.2             & {\bf 52.0}          &    {\bf 37.7}          & {\bf 39.4}                \\
     & \multirow{1}{*}{LCA=2}                     &      89.5             &   87.7         &   {\bf 72.5}           &      {\bf 31.5}             &   {\bf 62.3}         &    {\bf 69.9}          &      16.9             &   {\bf 56.3}         &    {\bf 38.8}          & {\bf 39.8}               \\
\Xhline{3\arrayrulewidth}
\multirow{2}{*}{TransFusion-L}    & \multirow{1}{*}{LCA=0}                     &      84.4             &   84.5         &    58.5           &      15.1             &   44.9         &    57.2          &      1.0             &   15.1         &    3.2          &      19.6              \\
\multirow{2}{*}{(LiDAR-only)}& \multirow{1}{*}{LCA=1}             &      85.5             &   85.7         &    67.4           &      21.8             &   46.7         &    59.1          &      1.6             &   21.8         &    3.7          &      19.8                \\
    & \multirow{1}{*}{LCA=2}                     &      85.5             &   86.1         &    67.5           &      22.6             &   47.7         &    59.9          &      1.7             &   22.6         &    4.2          &      20.4              \\
\midrule
    \multirow{2}{*}{TransFusion}  & \multirow{1}{*}{LCA=0}                    &      84.4     &   84.2      &    58.4           &      24.5          &     46.7       &    60.8       &      3.1           &   21.6         &    13.3     &    25.3    \\ 
    \multirow{2}{*}{(LiDAR + RGB)}   & \multirow{1}{*}{LCA=1}              &      86.0     &   85.4      &    67.3           &      26.3          &     50.1       &    63.5       &      14.4           &   34.7         &    20.6     &    25.6    \\
     & \multirow{1}{*}{LCA=2}                                                 &      86.0     &   85.9      &    67.4           &      26.8          &     52.2       &    65.1       &      15.2           &   36.1         &    22.8     &    26.4    \\
    \midrule
\multirow{2}{*}{TransFusion}    & \multirow{1}{*}{LCA=0}                     &      84.4             &   84.2         &    58.4           &      25.3             &   52.3         &    62.8          &      4.0              &   27.5         &    14.7          &      27.3              \\
 \multirow{2}{*}{\em w/ Multimodal Filtering} & \multirow{1}{*}{LCA=1}                     &      86.0             &   85.4         &    67.3           &      26.6             &   55.7         &    64.0          &      {\bf 25.1}             &   46.7         &    24.3          &      27.4                \\
    & \multirow{1}{*}{LCA=2}                     &      86.0             &   85.9         &    67.4           &      27.0             &   56.9         &    64.3          &     {\bf 25.8}             &   48.6         &    28.3          &      27.9              \\
\bottomrule
\end{tabular}
\label{tab:ap_h}
\end{table}
}

\begin{figure*}[h]
\centering
\hspace{-5mm} {\tt Vehicle} \hspace{27mm} {\tt Pedestrian} \hspace{27mm} {\tt Movable}\\
\includegraphics[width=0.32\linewidth,valign=t]{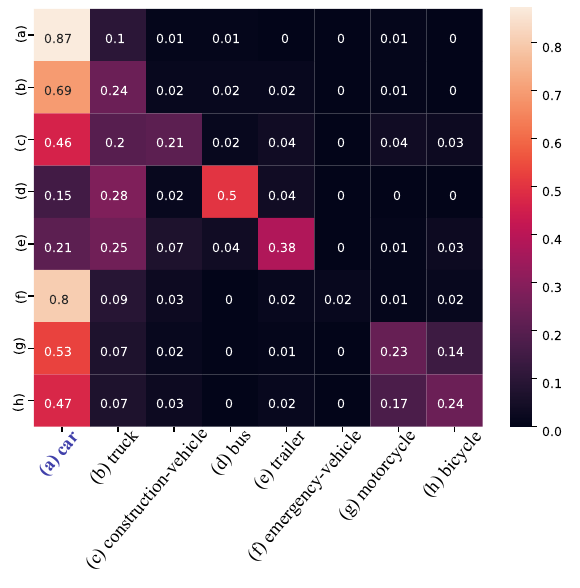} \hfill
\includegraphics[width=0.32\linewidth,valign=t]{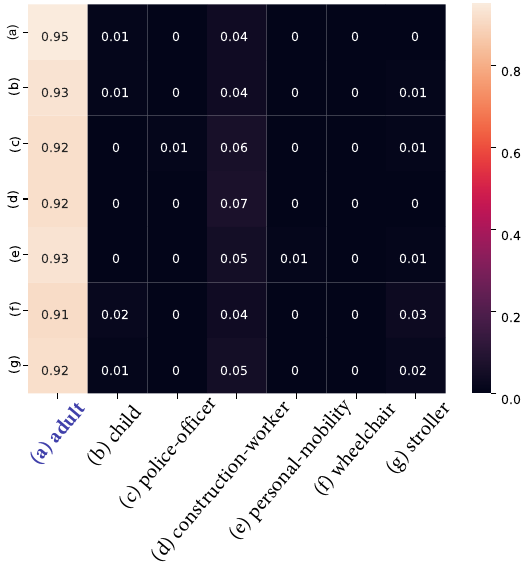} \hfill
\includegraphics[width=0.32\linewidth,valign=t]{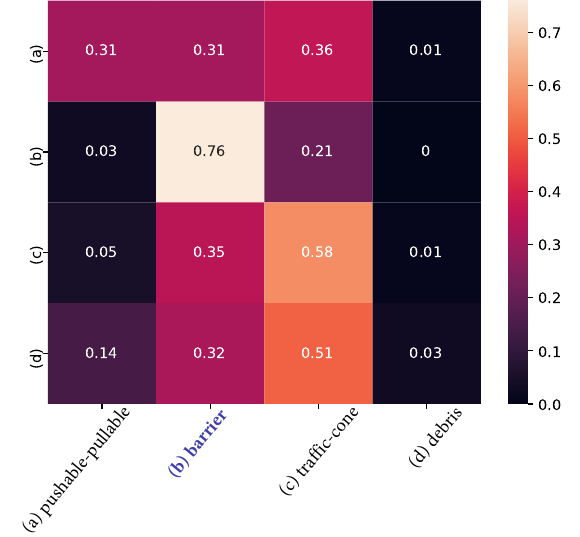}\hfill
\caption{\small 
{\bf Breakdown analysis of misclassifications within superclasses.} We analyze our best-performing model (CenterPoint w/ hierarchical training and multimodal filtering). 
Fine-grained classes are most often confused by the dominant class (in {\bf \color{blue}blue}) in each superclass: ({\bf left}) {\tt Vehicle} is dominated by {\tt car}, ({\bf mid}) {\tt Pedestrian} is dominated by {\tt adult}, and ({\bf right}) {\tt Movable} is dominated by {\tt barrier}.
We find that class confusions are reasonable. {\tt Car} is often mistaken for {\tt truck}. Similarly, {\tt truck}, {\tt construction-vehicle} and {\tt emergency-vehicle} are most often mistaken for {\tt car}. 
{\tt Bicycle} and {\tt motorcycle} are sometimes misclassified as {\tt car}, presumably because they are sometimes spatially close (within the 2m match threshold) to {\tt cars}.
{\tt Adults} have similar appearance to {\tt police-officer} and {\tt construction-worker}, and they are often co-localized with {\tt child} and {\tt stroller}; all of these might cause significant class confusion.
}

\vspace{-2mm}
\label{fig:conf_matrix}
\end{figure*}

\textbf{Analysis of Misclassifications}. For 3D detection, localization and classification are two important measures of 3D detection performance. In practice, we cannot achieve perfect performance for either. 
In safety-critical applications, detecting but misclassifying objects (as a semantically related category) is more desirable than a missed detection (e.g., detect but misclassify a {\tt child} as {\tt adult} versus not detecting this {\tt child}). 
Therefore, we introduce hierarchical AP (AP$_H$), which considers such semantic relationships across classes to award partial credit. Applying this hierarchical AP reveals that classes are most often misclassified as their LCA=1 siblings within coarse-grained superclasses.
We use confusion matrices to further analyze the misclassifications within superclasses, as shown in Fig.~\ref{fig:conf_matrix}. 
Below, we explain how to compute a confusion matrix for the detection task.

For each superclass, we make a confusion matrix, in which the entry $(i,j)$ indicates the misclassification rate of class-$i$ objects as class-$j$.
Specifically, given a fine-grained class $i$, we find its predictions that match ground-truth boxes within 2m center-distance of class-$i$ and all its sibling classes (LCA=1, within the corresponding superclass); we ignore all unmatched detections.
This allows us to count the mis-classifications of class-$i$ objects into class-$j$, with which a simple normalization produces misclassification rates. 

\textbf{Limitations.}
\label{sec:limitation}
\problem \ emphasizes object detection for {\tt rare} classes which can be safety-critical for downstream AV tasks such as motion planning and collision avoidance. However, our work does not study how solving \problem\ directly affects these tasks. Future work should address this limitation.
Another limitation, shared by contemporary benchmarks, is that our setup does not consider the correlation between individual classes. For example, 
the rare-class {\tt stroller} is often pushed by an {\tt adult}. One may argue that detecting {\tt adult} is sufficient for safe navigation. However, edge cases can occur in the real world where a {\tt stroller} can be unattended. 
\vspace{-1mm}
\section{Conclusion}
We explore the problem of long-tailed 3D detection (\problem), detecting  objects not only from {\tt common} classes but also from many {\tt rare} classes.
This problem is motivated by the operational safety of autonomous vehicles (AVs), but has broad robotic applications, 
e.g., elder-assistive robots \cite{savage2022elderly} that fetch diverse items \cite{grauman2022ego4d} should address \problem. To study \problem, we establish rigorous evaluation protocols that allow for partial credit to better diagnose 3D detectors.
We propose several algorithmic innovations to improve \problem, including a group-free detector head, hierarchical losses that promote feature sharing across long-tailed classes, and a simple multimodal fusion method that uses RGB-based detections to filter LiDAR-based detections, achieving significant improvement for \problem.


\section*{Acknowledgement}
This work was supported by the CMU Argo AI Center for Autonomous Vehicle Research.

\bibliography{references}  

\begin{thebibliography}{59}
\providecommand{\natexlab}[1]{#1}
\providecommand{\url}[1]{\texttt{#1}}
\expandafter\ifx\csname urlstyle\endcsname\relax
  \providecommand{\doi}[1]{doi: #1}\else
  \providecommand{\doi}{doi: \begingroup \urlstyle{rm}\Url}\fi

\bibitem[Geiger et~al.(2012)Geiger, Lenz, and Urtasun]{geiger2012we}
A.~Geiger, P.~Lenz, and R.~Urtasun.
\newblock Are we ready for autonomous driving? the kitti vision benchmark
  suite.
\newblock In \emph{IEEE Conference on Computer Vision and Pattern Recognition},
  2012.

\bibitem[Caesar et~al.(2020)Caesar, Bankiti, Lang, Vora, Liong, Xu, Krishnan,
  Pan, Baldan, and Beijbom]{caesar2020nuscenes}
H.~Caesar, V.~Bankiti, A.~H. Lang, S.~Vora, V.~E. Liong, Q.~Xu, A.~Krishnan,
  Y.~Pan, G.~Baldan, and O.~Beijbom.
\newblock nuscenes: A multimodal dataset for autonomous driving.
\newblock In \emph{Proceedings of the IEEE/CVF Conference on Computer Vision
  and Pattern Recognition}, 2020.

\bibitem[Chang et~al.(2019)Chang, Lambert, Sangkloy, Singh, Bak, Hartnett,
  Wang, Carr, Lucey, Ramanan, et~al.]{chang2019argoverse}
M.-F. Chang, J.~Lambert, P.~Sangkloy, J.~Singh, S.~Bak, A.~Hartnett, D.~Wang,
  P.~Carr, S.~Lucey, D.~Ramanan, et~al.
\newblock Argoverse: 3d tracking and forecasting with rich maps.
\newblock In \emph{IEEE/CVF Conference on Computer Vision and Pattern
  Recognition}, 2019.

\bibitem[Sun et~al.(2020)Sun, Kretzschmar, Dotiwalla, Chouard, Patnaik, Tsui,
  Guo, Zhou, Chai, Caine, Vasudevan, Han, Ngiam, Zhao, Timofeev, Ettinger,
  Krivokon, Gao, Joshi, Zhang, Shlens, Chen, and Anguelov]{sun2020waymo}
P.~Sun, H.~Kretzschmar, X.~Dotiwalla, A.~Chouard, V.~Patnaik, P.~Tsui, J.~Guo,
  Y.~Zhou, Y.~Chai, B.~Caine, V.~Vasudevan, W.~Han, J.~Ngiam, H.~Zhao,
  A.~Timofeev, S.~Ettinger, M.~Krivokon, A.~Gao, A.~Joshi, Y.~Zhang, J.~Shlens,
  Z.~Chen, and D.~Anguelov.
\newblock Scalability in perception for autonomous driving: Waymo open dataset.
\newblock In \emph{IEEE/CVF Conference on Computer Vision and Pattern
  Recognition (CVPR)}, 2020.

\bibitem[Taeihagh and Lim(2019)]{taeihagh2019governing}
A.~Taeihagh and H.~S.~M. Lim.
\newblock Governing autonomous vehicles: emerging responses for safety,
  liability, privacy, cybersecurity, and industry risks.
\newblock \emph{Transport Reviews}, 39\penalty0 (1):\penalty0 103--128, 2019.

\bibitem[Wong et~al.(2020)Wong, Wang, Ren, Liang, and
  Urtasun]{wong2020identifying}
K.~Wong, S.~Wang, M.~Ren, M.~Liang, and R.~Urtasun.
\newblock Identifying unknown instances for autonomous driving.
\newblock In \emph{CoRL}, 2020.

\bibitem[Liu et~al.(2019)Liu, Miao, Zhan, Wang, Gong, and
  Yu]{openlongtailrecognition}
Z.~Liu, Z.~Miao, X.~Zhan, J.~Wang, B.~Gong, and S.~X. Yu.
\newblock Large-scale long-tailed recognition in an open world.
\newblock In \emph{IEEE Conference on Computer Vision and Pattern Recognition
  (CVPR)}, 2019.

\bibitem[Lang et~al.(2019)Lang, Vora, Caesar, Zhou, Yang, and
  Beijbom]{lang2019pointpillars}
A.~H. Lang, S.~Vora, H.~Caesar, L.~Zhou, J.~Yang, and O.~Beijbom.
\newblock Pointpillars: Fast encoders for object detection from point clouds.
\newblock In \emph{{IEEE} Conference on Computer Vision and Pattern Recognition
  (CVPR)}, 2019.

\bibitem[Zhu et~al.(2019)Zhu, Jiang, Zhou, Li, and Yu]{zhu2019class}
B.~Zhu, Z.~Jiang, X.~Zhou, Z.~Li, and G.~Yu.
\newblock Class-balanced grouping and sampling for point cloud 3d object
  detection.
\newblock \emph{arXiv preprint arXiv:1908.09492}, 2019.

\bibitem[Hu et~al.(2019)Hu, Ziglar, Held, and Ramanan]{hu2019wysiwyg}
P.~Hu, J.~Ziglar, D.~Held, and D.~Ramanan.
\newblock What you see is what you get: Exploiting visibility for 3d object
  detection.
\newblock \emph{CoRR}, abs/1912.04986, 2019.

\bibitem[Yan et~al.(2018)Yan, Mao, and Li]{yan2018second}
Y.~Yan, Y.~Mao, and B.~Li.
\newblock Second: Sparsely embedded convolutional detection.
\newblock \emph{Sensors}, 18\penalty0 (10):\penalty0 3337, 2018.

\bibitem[Wang et~al.(2019)Wang, Wang, Zhang, Gu, and Hwang]{wang2019exploit}
G.~Wang, Y.~Wang, H.~Zhang, R.~Gu, and J.-N. Hwang.
\newblock Exploit the connectivity: Multi-object tracking with trackletnet.
\newblock In \emph{Proceedings of the 27th ACM International Conference on
  Multimedia}, pages 482--490, 2019.

\bibitem[Yin et~al.(2020)Yin, Zhou, and Kr{\"a}henb{\"u}hl]{yin2020center}
T.~Yin, X.~Zhou, and P.~Kr{\"a}henb{\"u}hl.
\newblock Center-based 3d object detection and tracking.
\newblock \emph{arXiv preprint arXiv:2006.11275}, 2020.

\bibitem[Chen et~al.(2017)Chen, Ma, Wan, Li, and Xia]{chen2017multiview}
X.~Chen, H.~Ma, J.~Wan, B.~Li, and T.~Xia.
\newblock Multi-view 3d object detection network for autonomous driving.
\newblock In \emph{Proceedings of the IEEE conference on Computer Vision and
  Pattern Recognition}, 2017.

\bibitem[Ku et~al.(2018)Ku, Mozifian, Lee, Harakeh, and Waslander]{ku2018joint}
J.~Ku, M.~Mozifian, J.~Lee, A.~Harakeh, and S.~Waslander.
\newblock Joint 3d proposal generation and object detection from view
  aggregation.
\newblock \emph{IROS}, 2018.

\bibitem[Yoo et~al.(2020)Yoo, Kim, Kim, and Choi]{yoo203dcvf}
J.~H. Yoo, Y.~Kim, J.~S. Kim, and J.~W. Choi.
\newblock 3d-cvf: Generating joint camera and lidar features using cross-view
  spatial feature fusion for 3d object detection.
\newblock In \emph{European Conference on Computer Vision (ECCV)}, 2020.

\bibitem[Bai et~al.(2022)Bai, Hu, Zhu, Huang, Chen, Fu, and
  Tai]{bai22transfusion}
X.~Bai, Z.~Hu, X.~Zhu, Q.~Huang, Y.~Chen, H.~Fu, and C.~Tai.
\newblock Transfusion: Robust lidar-camera fusion for 3d object detection with
  transformers.
\newblock \emph{CoRR}, abs/2203.11496, 2022.

\bibitem[Chen et~al.(2022)Chen, Shi, Ye, Mertz, Ramanan, and
  Kong]{chen2021multimodal}
Y.-T. Chen, J.~Shi, Z.~Ye, C.~Mertz, D.~Ramanan, and S.~Kong.
\newblock Multimodal object detection via probabilistic ensembling.
\newblock In \emph{European Conference on Computer Vision (ECCV)}, 2022.

\bibitem[Gupta et~al.(2022)Gupta, Kanjani, Li, Ferroni, Hays, Ramanan, and
  Kong]{gupta2023far3det}
S.~Gupta, J.~Kanjani, M.~Li, F.~Ferroni, J.~Hays, D.~Ramanan, and S.~Kong.
\newblock Far3det: Towards far-field 3d detection.
\newblock In \emph{NeurIPS}, 2022.

\bibitem[Sindagi et~al.(2019)Sindagi, Zhou, and Tuzel]{sindagi2019mvxnet}
V.~A. Sindagi, Y.~Zhou, and O.~Tuzel.
\newblock Mvx-net: Multimodal voxelnet for 3d object detection.
\newblock In \emph{International Conference on Robotics and Automation (ICRA)},
  2019.

\bibitem[You et~al.(2020)You, Wang, Chao, Garg, Pleiss, Hariharan, Campbell,
  and Weinberger]{you2020pseudoLiDAR}
Y.~You, Y.~Wang, W.-L. Chao, D.~Garg, G.~Pleiss, B.~Hariharan, M.~Campbell, and
  K.~Q. Weinberger.
\newblock Pseudo-lidar++: Accurate depth for 3d object detection in autonomous
  driving.
\newblock \emph{arXiv arXiv:1906.06310}, 2020.

\bibitem[Vora et~al.(2020)Vora, Lang, Helou, and
  Beijbom]{vora2020pointpainting}
S.~Vora, A.~H. Lang, B.~Helou, and O.~Beijbom.
\newblock Pointpainting: Sequential fusion for 3d object detection.
\newblock In \emph{IEEE/CVF conference on computer vision and pattern
  recognition}, 2020.

\bibitem[Yin et~al.(2021)Yin, Zhou, and Kr{\"a}henb{\"u}hl]{yin2021multimodal}
T.~Yin, X.~Zhou, and P.~Kr{\"a}henb{\"u}hl.
\newblock Multimodal virtual point 3d detection.
\newblock \emph{NeurIPS}, 2021.

\bibitem[Qi et~al.(2018)Qi, Liu, Wu, Su, and Guibas]{qi2018frustum}
C.~R. Qi, W.~Liu, C.~Wu, H.~Su, and L.~J. Guibas.
\newblock Frustum pointnets for 3d object detection from rgb-d data.
\newblock In \emph{IEEE conference on computer vision and pattern recognition},
  2018.

\bibitem[Xu et~al.(2018)Xu, Anguelov, and Jain]{xu2018pointfusion}
D.~Xu, D.~Anguelov, and A.~Jain.
\newblock Pointfusion: Deep sensor fusion for 3d bounding box estimation.
\newblock In \emph{IEEE conference on computer vision and pattern recognition},
  2018.

\bibitem[Pang et~al.(2020)Pang, Morris, and Radha]{pang2009clocs}
S.~Pang, D.~Morris, and H.~Radha.
\newblock Clocs: Camera-lidar object candidates fusion for 3d object detection.
\newblock In \emph{IEEE/RSJ International Conference on Intelligent Robots and
  Systems (IROS)}, 2020.

\bibitem[Reed(2001)]{reed2001pareto}
W.~J. Reed.
\newblock The pareto, zipf and other power laws.
\newblock \emph{Economics letters}, 74\penalty0 (1):\penalty0 15--19, 2001.

\bibitem[Zhang et~al.(2021)Zhang, Kang, Hooi, Yan, and Feng]{zhang2021deep}
Y.~Zhang, B.~Kang, B.~Hooi, S.~Yan, and J.~Feng.
\newblock Deep long-tailed learning: A survey.
\newblock \emph{arXiv:2110.04596}, 2021.

\bibitem[Alshammari et~al.(2022)Alshammari, Wang, Ramanan, and
  Kong]{alshammari2022long}
S.~Alshammari, Y.-X. Wang, D.~Ramanan, and S.~Kong.
\newblock Long-tailed recognition via weight balancing.
\newblock In \emph{IEEE/CVF Conference on Computer Vision and Pattern
  Recognition}, 2022.

\bibitem[Cui et~al.(2019)Cui, Jia, Lin, Song, and Belongie]{cui2019class}
Y.~Cui, M.~Jia, T.-Y. Lin, Y.~Song, and S.~Belongie.
\newblock Class-balanced loss based on effective number of samples.
\newblock In \emph{CVPR}, 2019.

\bibitem[Khan et~al.(2017)Khan, Hayat, Bennamoun, Sohel, and
  Togneri]{khan2017cost}
S.~H. Khan, M.~Hayat, M.~Bennamoun, F.~A. Sohel, and R.~Togneri.
\newblock Cost-sensitive learning of deep feature representations from
  imbalanced data.
\newblock \emph{IEEE transactions on neural networks and learning systems},
  29\penalty0 (8):\penalty0 3573--3587, 2017.

\bibitem[Cao et~al.(2019)Cao, Wei, Gaidon, Arechiga, and Ma]{cao2019learning}
K.~Cao, C.~Wei, A.~Gaidon, N.~Arechiga, and T.~Ma.
\newblock Learning imbalanced datasets with label-distribution-aware margin
  loss.
\newblock In \emph{NeurIPS}, 2019.

\bibitem[Khan et~al.(2019)Khan, Hayat, Zamir, Shen, and Shao]{khan2019striking}
S.~Khan, M.~Hayat, S.~W. Zamir, J.~Shen, and L.~Shao.
\newblock Striking the right balance with uncertainty.
\newblock In \emph{CVPR}, 2019.

\bibitem[Huang et~al.(2019)Huang, Li, Loy, and Tang]{huang2019deep}
C.~Huang, Y.~Li, C.~C. Loy, and X.~Tang.
\newblock Deep imbalanced learning for face recognition and attribute
  prediction.
\newblock \emph{PAMI}, 42\penalty0 (11):\penalty0 2781--2794, 2019.

\bibitem[Zhang et~al.(2021)Zhang, Li, Yan, He, and Sun]{zhang2021distribution}
S.~Zhang, Z.~Li, S.~Yan, X.~He, and J.~Sun.
\newblock Distribution alignment: A unified framework for long-tail visual
  recognition.
\newblock In \emph{CVPR}, 2021.

\bibitem[Drummond et~al.(2003)Drummond, Holte, et~al.]{drummond2003c4}
C.~Drummond, R.~C. Holte, et~al.
\newblock C4. 5, class imbalance, and cost sensitivity: why under-sampling
  beats over-sampling.
\newblock In \emph{Workshop on learning from imbalanced datasets II}, 2003.

\bibitem[Chawla et~al.(2002)Chawla, Bowyer, Hall, and
  Kegelmeyer]{chawla2002smote}
N.~V. Chawla, K.~W. Bowyer, L.~O. Hall, and W.~P. Kegelmeyer.
\newblock Smote: synthetic minority over-sampling technique.
\newblock \emph{Journal of Artificial Intelligence Research}, 16:\penalty0
  321--357, 2002.

\bibitem[Han et~al.(2005)Han, Wang, and Mao]{han2005borderline}
H.~Han, W.-Y. Wang, and B.-H. Mao.
\newblock Borderline-smote: a new over-sampling method in imbalanced data sets
  learning.
\newblock In \emph{International Conference on Intelligent Computing}, pages
  878--887. Springer, 2005.

\bibitem[Tang et~al.(2020)Tang, Huang, and Zhang]{tang2020long}
K.~Tang, J.~Huang, and H.~Zhang.
\newblock Long-tailed classification by keeping the good and removing the bad
  momentum causal effect.
\newblock In \emph{NeurIPS}, 2020.

\bibitem[Gupta et~al.(2019)Gupta, Dollar, and Girshick]{gupta2019lvis}
A.~Gupta, P.~Dollar, and R.~Girshick.
\newblock Lvis: A dataset for large vocabulary instance segmentation.
\newblock In \emph{CVPR}, 2019.

\bibitem[Wang et~al.(2021)Wang, Zhu, Pang, and Lin]{wang2021fcos3d}
T.~Wang, X.~Zhu, J.~Pang, and D.~Lin.
\newblock Fcos3d: Fully convolutional one-stage monocular 3d object detection.
\newblock In \emph{IEEE/CVF International Conference on Computer Vision}, pages
  913--922, 2021.

\bibitem[Everingham et~al.(2015)Everingham, Eslami, Van~Gool, Williams, Winn,
  and Zisserman]{everingham2015pascal}
M.~Everingham, S.~A. Eslami, L.~Van~Gool, C.~K. Williams, J.~Winn, and
  A.~Zisserman.
\newblock The pascal visual object classes challenge: A retrospective.
\newblock \emph{International Journal of Computer Vision (IJCV)}, 2015.

\bibitem[Lin et~al.(2014)Lin, Maire, Belongie, Hays, Perona, Ramanan,
  Doll{\'{a}}r, and Zitnick]{lin2014coco}
T.~Lin, M.~Maire, S.~J. Belongie, J.~Hays, P.~Perona, D.~Ramanan,
  P.~Doll{\'{a}}r, and C.~L. Zitnick.
\newblock Microsoft {COCO:} common objects in context.
\newblock In \emph{European Conference on Computer Vision (ECCV)}, 2014.

\bibitem[Russakovsky et~al.(2015)Russakovsky, Deng, Su, Krause, Satheesh, Ma,
  Huang, Karpathy, Khosla, Bernstein, et~al.]{russakovsky2015imagenet}
O.~Russakovsky, J.~Deng, H.~Su, J.~Krause, S.~Satheesh, S.~Ma, Z.~Huang,
  A.~Karpathy, A.~Khosla, M.~Bernstein, et~al.
\newblock Imagenet large scale visual recognition challenge.
\newblock \emph{International journal of computer vision}, 115\penalty0
  (3):\penalty0 211--252, 2015.

\bibitem[Carion et~al.(2020)Carion, Massa, Synnaeve, Usunier, Kirillov, and
  Zagoruyko]{detr2020carion}
N.~Carion, F.~Massa, G.~Synnaeve, N.~Usunier, A.~Kirillov, and S.~Zagoruyko.
\newblock End-to-end object detection with transformers.
\newblock In \emph{European Conference on Computer Vision (ECCV)}, 2020.

\bibitem[Paszke et~al.(2019)Paszke, Gross, Massa, Lerer, Bradbury, Chanan,
  Killeen, Lin, Gimelshein, Antiga, Desmaison, K{\"{o}}pf, Yang, DeVito,
  Raison, Tejani, Chilamkurthy, Steiner, Fang, Bai, and
  Chintala]{paszke2019pytorch}
A.~Paszke, S.~Gross, F.~Massa, A.~Lerer, J.~Bradbury, G.~Chanan, T.~Killeen,
  Z.~Lin, N.~Gimelshein, L.~Antiga, A.~Desmaison, A.~K{\"{o}}pf, E.~Yang,
  Z.~DeVito, M.~Raison, A.~Tejani, S.~Chilamkurthy, B.~Steiner, L.~Fang,
  J.~Bai, and S.~Chintala.
\newblock Pytorch: An imperative style, high-performance deep learning library.
\newblock In \emph{Advances in Neural Information Processing Systems}, 2019.

\bibitem[Kingma and Ba(2015)]{kingma2015adam}
D.~P. Kingma and J.~Ba.
\newblock Adam: {A} method for stochastic optimization.
\newblock In \emph{International Conference on Learning Representations
  (ICLR)}, 2015.

\bibitem[Smith(2017)]{smith2017cyclical}
L.~N. Smith.
\newblock Cyclical learning rates for training neural networks.
\newblock In \emph{IEEE Winter Conference on Applications of Computer Vision,
  {WACV}}, 2017.

\bibitem[Wilson et~al.(2021)Wilson, Qi, Agarwal, Lambert, Singh, Khandelwal,
  Pan, Kumar, Hartnett, Pontes, Ramanan, Carr, and Hays]{wilson2021argoverse}
B.~Wilson, W.~Qi, T.~Agarwal, J.~Lambert, J.~Singh, S.~Khandelwal, B.~Pan,
  R.~Kumar, A.~Hartnett, J.~K. Pontes, D.~Ramanan, P.~Carr, and J.~Hays.
\newblock Argoverse 2: Next generation datasets for self-driving perception and
  forecasting.
\newblock In \emph{Neural Information Processing Systems Datasets and
  Benchmarks Track}, 2021.

\bibitem[Van~Horn et~al.(2018)Van~Horn, Mac~Aodha, Song, Cui, Sun, Shepard,
  Adam, Perona, and Belongie]{van2018inaturalist}
G.~Van~Horn, O.~Mac~Aodha, Y.~Song, Y.~Cui, C.~Sun, A.~Shepard, H.~Adam,
  P.~Perona, and S.~Belongie.
\newblock The inaturalist species classification and detection dataset.
\newblock In \emph{CVPR}, 2018.

\bibitem[Liu et~al.(2019)Liu, Miao, Zhan, Wang, Gong, and Yu]{liu2019large}
Z.~Liu, Z.~Miao, X.~Zhan, J.~Wang, B.~Gong, and S.~X. Yu.
\newblock Large-scale long-tailed recognition in an open world.
\newblock In \emph{CVPR}, pages 2537--2546, 2019.

\bibitem[Savage(2022)]{savage2022elderly}
N.~Savage.
\newblock Robots rise to meet the challenge of caring for old people.
\newblock \emph{Nature}, 2022.

\bibitem[Grauman et~al.()Grauman, Westbury, Byrne, Chavis, Furnari, Girdhar,
  Hamburger, Jiang, Liu, Liu, Martin, Nagarajan, Radosavovic, Ramakrishnan,
  Ryan, Sharma, Wray, Xu, Xu, Zhao, Bansal, Batra, Cartillier, Crane, Do,
  Doulaty, Erapalli, Feichtenhofer, Fragomeni, Fu, Fuegen, Gebreselasie,
  Gonzalez, Hillis, Huang, Huang, Jia, Khoo, Kolar, Kottur, Kumar, Landini, Li,
  Li, Li, Mangalam, Modhugu, Munro, Murrell, Nishiyasu, Price, Puentes,
  Ramazanova, Sari, Somasundaram, Southerland, Sugano, Tao, Vo, Wang, Wu, Yagi,
  Zhu, Arbelaez, Crandall, Damen, Farinella, Ghanem, Ithapu, Jawahar, Joo,
  Kitani, Li, Newcombe, Oliva, Park, Rehg, Sato, Shi, Shou, Torralba,
  Torresani, Yan, and Malik]{grauman2022ego4d}
K.~Grauman, A.~Westbury, E.~Byrne, Z.~Chavis, A.~Furnari, R.~Girdhar,
  J.~Hamburger, H.~Jiang, M.~Liu, X.~Liu, M.~Martin, T.~Nagarajan,
  I.~Radosavovic, S.~K. Ramakrishnan, F.~Ryan, J.~Sharma, M.~Wray, M.~Xu, E.~Z.
  Xu, C.~Zhao, S.~Bansal, D.~Batra, V.~Cartillier, S.~Crane, T.~Do, M.~Doulaty,
  A.~Erapalli, C.~Feichtenhofer, A.~Fragomeni, Q.~Fu, C.~Fuegen,
  A.~Gebreselasie, C.~Gonzalez, J.~Hillis, X.~Huang, Y.~Huang, W.~Jia, W.~Khoo,
  J.~Kolar, S.~Kottur, A.~Kumar, F.~Landini, C.~Li, Y.~Li, Z.~Li, K.~Mangalam,
  R.~Modhugu, J.~Munro, T.~Murrell, T.~Nishiyasu, W.~Price, P.~R. Puentes,
  M.~Ramazanova, L.~Sari, K.~Somasundaram, A.~Southerland, Y.~Sugano, R.~Tao,
  M.~Vo, Y.~Wang, X.~Wu, T.~Yagi, Y.~Zhu, P.~Arbelaez, D.~Crandall, D.~Damen,
  G.~M. Farinella, B.~Ghanem, V.~K. Ithapu, C.~V. Jawahar, H.~Joo, K.~Kitani,
  H.~Li, R.~Newcombe, A.~Oliva, H.~S. Park, J.~M. Rehg, Y.~Sato, J.~Shi, M.~Z.
  Shou, A.~Torralba, L.~Torresani, M.~Yan, and J.~Malik.
\newblock Ego4d: Around the world in 3, 000 hours of egocentric video.
\newblock \emph{Computer Vision and Pattern Recognition 2022}.

\bibitem[Lin et~al.(2017)Lin, Goyal, Girshick, He, and
  Doll{\'a}r]{lin2017focal}
T.-Y. Lin, P.~Goyal, R.~Girshick, K.~He, and P.~Doll{\'a}r.
\newblock Focal loss for dense object detection.
\newblock In \emph{ICCV}, 2017.

\bibitem[Ren et~al.(2015)Ren, He, Girshick, and Sun]{ren2015faster}
S.~Ren, K.~He, R.~Girshick, and J.~Sun.
\newblock Faster r-cnn: Towards real-time object detection with region proposal
  networks.
\newblock In \emph{Advances in neural information processing systems}, 2015.

\bibitem[Wu et~al.(2019)Wu, Tygert, and LeCun]{Wu2019AHL}
C.~J. Wu, M.~Tygert, and Y.~LeCun.
\newblock A hierarchical loss and its problems when classifying
  non-hierarchically.
\newblock \emph{PLoS ONE}, 14, 2019.

\bibitem[Li et~al.(2020)Li, Wang, Kang, Tang, Wang, Li, and
  Feng]{li2020classifierimbalance}
Y.~Li, T.~Wang, B.~Kang, S.~Tang, C.~Wang, J.~Li, and J.~Feng.
\newblock Overcoming classifier imbalance for long-tail object detection with
  balanced group softmax.
\newblock In \emph{2020 IEEE/CVF Conference on Computer Vision and Pattern
  Recognition (CVPR)}, 2020.

\bibitem[Welchman et~al.(2004)Welchman, Tuck, and Harris]{welchman2004human}
A.~E. Welchman, V.~L. Tuck, and J.~M. Harris.
\newblock Human observers are biased in judging the angular approach of a
  projectile.
\newblock \emph{Vision research}, 44\penalty0 (17):\penalty0 2027--2042, 2004.

\bibitem[Wang et~al.(2021)Wang, Zhu, Pang, and Lin]{wang21_fcos3d}
T.~Wang, X.~Zhu, J.~Pang, and D.~Lin.
\newblock {FCOS3D:} fully convolutional one-stage monocular 3d object
  detection.
\newblock In \emph{{IEEE/CVF} International Conference on Computer Vision
  Workshops}, 2021.

\end{thebibliography}

\appendix 

\begin{center}
\Large  {\bf Appendix}
\end{center}

\section{Implementation Details}
We follow the training procedure of the respective detectors \cite{lang2019pointpillars, yin2020center, bai22transfusion} which have open-source code. We describe important implementation details below.

\begin{itemize}
    \item {\em Model Architecture.} We adopt the architecture in \cite{zhu2019class} but make an important modification. The original architecture (for the standard nuScenes benchmark) has six heads designed for ten classes; each head has 64 filters. We first adapted this architecture for LT3D using seven heads designed for 18 classes. We then replace these seven heads with a single head consisting of 512 filters shared by all classes. 
    \item {\em Training Losses.} We use the sigmoid focal loss (for recognition) \cite{lin2017focal} and L1 regression loss (for localization) below. Existing works also use the same losses but only with fine labels; we apply the loss to both coarse and fine labels. Concretely, our loss function for CenterPoint is as follows: $L = L_{HM} + \lambda L_{REG}$, where $L_{HM} = \sum_{i=0}^{C} SigmoidFocalLoss(X_i, Y_i)$ and $L_{REG} = |X_{BOX} - Y_{BOX}|$, where $X_i$ and $Y_i$ are the $i^{th}$ class' predicted and ground-truth heat maps, while $X_{BOX}$ and $Y_{BOX}$ are the predicted and ground-truth box attributes. 
    Without our hierarchical loss, $C$=18. With our hierarchical loss, $C$=22 (18 fine grained + 3 coarse + 1 object class). $\lambda$ is set to 0.25. Modifications for other detectors similarly follow.
    \item {\em Post-processing.} 
    We use non-maximum suppression (NMS) on detections {\em within} each class to suppress lower-scoring detections. In contrast, existing works apply NMS on all detections {\em across} classes, i.e., suppressing detections overlapping other classes' detections (e.g., a pedestrian detection can suppress other pedestrian and traffic cone detections).
\end{itemize}

\section{Analysis on Multi-Head Architectures and Class Grouping}
Many contemporary networks use a multi-head architecture that groups classes of similar size and shape to facilitate efficient feature sharing. For example, CenterPoint groups {\tt pedestrian} and {\tt traffic-cone} since these objects are both tall and skinny. We study the impact of grouping for both the standard and LT3D problem setups. We define the groups used for this study below. Each group is enclosed in curly braces. Our group-free head includes all classes into a single group.

\begin{itemize}
    \item Original: \{{\tt Car}\}, \{{\tt Truck}, {\tt Construction Vehicle}\}, \{{\tt Bus}, {\tt Trailer}\}, \{{\tt Barrier}\}, \{{\tt Motorcycle}, {\tt Bicycle}\}, \{{\tt Pedestrian}, {\tt Traffic Cone}\}
    \item LT3D: \{{\tt Car}\}, \{{\tt Truck}, {\tt Construction Vehicle}\}, \{{\tt Bus}, {\tt Trailer}\}, \{{\tt Barrier}\}, \{{\tt Motorcycle}, {\tt Bicycle}\}, \{{\tt Adult}, {\tt Child}, {\tt Construction Worker}, {\tt Police Officer}, {\tt Traffic Cone}\}, \{{\tt Pushable Pullable}, {\tt Debris}, {\tt Stroller}, {\tt Personal Mobility}, {\tt Emergency Vehicle}\}
\end{itemize}

We use the class groups proposed by prior works \cite{zhu2019class, yin2020center} for the standard benchmark and adapt this grouping for \problem. Our proposed group-free detector head architecture consistently outperforms grouping-based approaches on both the standard and \problem \ benchmarks. We note that sub-optimal grouping strategies (such as those adopted for \problem) may yield significantly diminished performance, whereas optimized grouping strategies (such as those adopted for the standard setup) have comparable performance to the group-free approach. The group-free approach simplifies architecture design, while also providing competitive performance.

Two insights allow us to train the group-free architecture. First, we make the group-free head proportionally larger to train more classes. The standard grouping setup contains 6 heads, each with 64 convolutional filters. Scaling up to the nearest power of two, our group-free head has 512 convolutional filters. Second, we do not perform between-class NMS. The standard setup performs NMS between classes in each group (e.g., since pedestrians and traffic cones are tall and skinny, the model should only predict that an object is either a traffic cone or a pedestrian). However, performing NMS between classes requires that confidence scores are calibrated, which is not the case. Moreover, for LT3D, score calibration becomes more important for {\tt rare} classes as these classes have lower confidence scores than {\tt common} classes on average, meaning that {\tt common} objects will likely suppress {\tt rare} objects within the same group. Our solution is to only perform within-class NMS, which is standard for 2D detectors \cite{ren2015faster}. 

{
\setlength{\tabcolsep}{0.25em} 
\begin{table}[t] 
\small
\centering
\caption{\small 
Our proposed group-free detector head architecture consistently outperforms grouping-based approaches on both the standard and \problem \ benchmarks. We note that sub-optimal grouping strategies (such as those adopted for \problem) may yield significantly diminished performance, whereas optimized grouping strategies (such as those adopted for the standard setup) have comparable performance to the group-free approach. Note, {\tt TC} is {\tt traffic-cone}, {\tt CV} is {\tt construction vehicle}, {\tt MC} is {\tt motorcycle}, {\tt PP} is {\tt pushable-pullable}, {\tt CW} is {\tt construction-worker}, and {\tt PO} is {\tt police-officer}. We highlight classes with {\tt Medium} and {\tt Few} examples per class in {\color{blue}blue}.
}
\begin{tabular}{l c c c c c c c c c c c c c}
\toprule
CenterPoint & \multicolumn{1}{c}{Multi-Head} & \multicolumn{1}{c}{Car} & \multicolumn{1}{c}{Ped.} & \multicolumn{1}{c}{Barrier} & \multicolumn{1}{c}{TC} &  \multicolumn{1}{c}{Truck} & \multicolumn{1}{c}{Bus} & \multicolumn{1}{c}{Trailer} &  \multicolumn{1}{c}{CV} & \multicolumn{1}{c}{MC} & \multicolumn{1}{c}{Bicycle}  \\ 
\midrule
Original                  & \checkmark &  87.7   &  87.7   &  70.7   &  74.0   &  63.6   & 72.7  &  \textbf{45.1}   &  \textbf{26.3}   &  64.7  &  47.9   \\ 
                             &            &  \textbf{89.1}   &  \textbf{88.4}   &  \textbf{70.8}   &  \textbf{74.3}   &  \textbf{64.8}   & \textbf{72.9}  &  42.0   &  25.7   &  \textbf{65.9}  &  \textbf{53.6}   \\ 
\midrule
\multirow{1}{*}{for LT3D}    & \checkmark &  82.4   &  ---   &  62.0   &  60.1   &  49.4   & 55.7  &  28.9   &  19.7   &  48.9  &  33.6   \\ 
                             &            &  \textbf{88.1}   &  ---   &  \textbf{72.4}   &  \textbf{72.7}   &  \textbf{62.7}   & \textbf{70.8}  &  \textbf{40.2}   &  \textbf{24.5}   &  \textbf{62.8}  &  \textbf{48.5}   \\ 
\midrule
\midrule 
\multirow{1}{*}{} & \multirow{1}{*}{}  & \multicolumn{1}{c}{Adult} & \multicolumn{1}{c}{\color{blue}PP} & \multicolumn{1}{c}{\color{blue}CW} & \multicolumn{1}{c}{\color{blue}Debris} &  \multicolumn{1}{c}{\color{blue}Child} & \multicolumn{1}{c}{\color{blue}Stroller} & \multicolumn{1}{c}{\color{blue}PO} &  \multicolumn{1}{c}{\color{blue}EV} & \multicolumn{1}{c}{\color{blue}PM} & \multicolumn{1}{c}{{\tt All}} \\ 
\midrule
Original                  & \checkmark &  ---   &  ---   &  ---   &  ---   &  ---   & ---  &  ---  &  ---  &  --- & 64.0 \\ 
                          &            &  ---   &  ---   &  ---   &  ---   &  ---   & ---  &  ---  &  ---   & --- & \textbf{64.8} \\ 
\midrule
\multirow{1}{*}{for LT3D}    & \checkmark &  81.2   &  21.7   &  14.2   & 1.1   &  \textbf{0.1}     & 0.1  &  \textbf{1.3}   &  0.1          &   \textbf{0.1}     & 31.2      \\ 
                             &            &  \textbf{86.3}   &  \textbf{32.7}   &  \textbf{22.2}   & \textbf{4.3}   &  \textbf{0.1}     & \textbf{4.3}  &  \textbf{1.8}    & 10.3          &   \textbf{0.1}  & \textbf{39.2}       \\ 
\bottomrule
\end{tabular}
\label{tab:segm}
\end{table}
}

\section{Ablation on Hierarchical Training and Inference}
Classic methods train a hierarchical softmax (in contrast to our simple approach of sigmoid focal loss with both fine and coarse classes), where one multiplies the class probabilities of the hierarchical predictions during training and inference \cite{Wu2019AHL}. We implemented such an approach, but found the training did not converge. Interestingly, \cite{Wu2019AHL} shows such a hierarchical softmax loss has little impact on long-tailed object detection (in 2D images), which is one reason they have not been historically adopted. Instead, we found better results using the method from \cite{li2020classifierimbalance} (a winning 2D object detection system on the LVIS \cite{gupta2019lvis} benchmark) which multiples class probabilities of predictions (e.g. $P_{CAR} = P_{OBJ} * P_{CAR}$) at test-time, even when such predictions are not trained with a hierarchical softmax. We tested three variants and compared it to our approach (which recall, uses only fine-grained class probabilities at inference). Table~\ref{tab:hierarchy_loss} compares their performance for \problem.
{
\setlength{\tabcolsep}{1.0em} 
\begin{table}[h]
\small
\centering
\caption{\small 
Different variants achieve similar performance. We note that other methods do improve accuracy in the tail by sacrificing performance in the head, suggesting that hybrid approaches that apply different techniques for head-vs-tail classes may further improve accuracy. Unlike \cite{li2020classifierimbalance, Wu2019AHL} which requires a strict label hierarchy, our approach is not limited to a hierarchy.}
\begin{tabular}{l c c c c c c c c c c c}
\toprule
\multirow{1}{*}{Method} & \multirow{1}{*}{Hierarchy} & \multicolumn{1}{c}{\tt Many} & \multicolumn{1}{c}{\texttt{Medium}} & \multicolumn{1}{c}{\texttt{Few}} & \multicolumn{1}{c}{\tt All} \\ 
\midrule
\multirow{1}{*}{CenterPoint (w/o Hierarchy) \cite{yin2020center}}                       &      n/a      &  76.4             & 43.1         &   3.5        &   39.2        \\ 
\midrule
\multirow{4}{*}{CenterPoint w/ Hierarchy}        &  (a)   & \textbf{77.1}             & {45.1}          &   4.3      &   40.4       \\ 
\multirow{1}{*}{}                                &  (b)   &  {76.4}             & {45.0}          &   5.3      &   40.5       \\ 
\multirow{1}{*}{}                                &  (c)   &  {76.5}             & \textbf{45.2}          &   5.2      &   \textbf{40.6}       \\
\multirow{1}{*}{}                                &  (d)   &  {74.5}             & {43.5}          &   \textbf{5.6}      &   39.5       \\ 
\bottomrule
\end{tabular}
\label{tab:hierarchy_loss}
\end{table}
}

\begin{enumerate}[label=(\alph*)]
    \item Ours (e.g., Finegrain score only)
    \item Object score * Finegrain score (\cite{li2020classifierimbalance}, e.g. $P_{CAR} = P_{OBJ} * P_{CAR}$) 
    \item Coarse score * Finegrain score (Variant-1 of \cite{li2020classifierimbalance}, e.g. $P_{CAR} = P_{VEHICLE} * P_{CAR}$)
    \item Object score * Coarse score * Finegrain score (Variant-2 of \cite{li2020classifierimbalance}, e.g. $P_{CAR} = P_{OBJ} * P_{VEHICLE} * P_{CAR}$)
\end{enumerate}

Unlike \cite{li2020classifierimbalance, Wu2019AHL} which require a strict label hierarchy, our approach is not limited to a hierarchy. We find that other hierarchical methods improve accuracy in the tail by sacrificing performance in the head, suggesting that hybrid approaches that apply different techniques for head-vs-tail classes may further improve accuracy.

\section{Evaluating on Argoverse 2.0}
 \begin{figure}[h]
\centering
\hspace{-64mm} {\tt Few} \hspace{10mm} {\tt Medium} \hspace{10mm} {\tt Many}\\
\includegraphics[width=0.48\linewidth, clip, trim={0cm 0cm 0cm 0cm}]{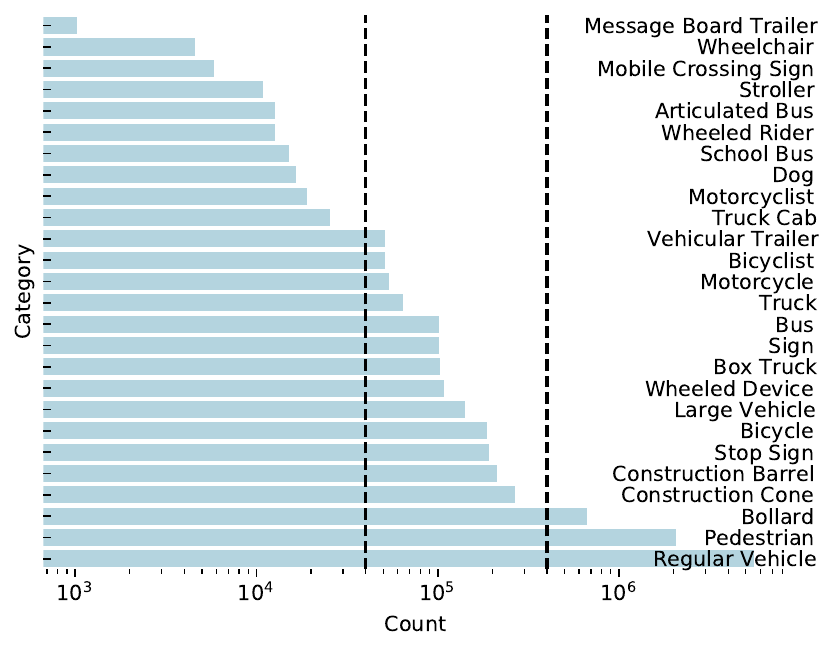} \hfill
\includegraphics[width=0.5\linewidth, clip, trim={4cm 0cm 3cm 0cm}]{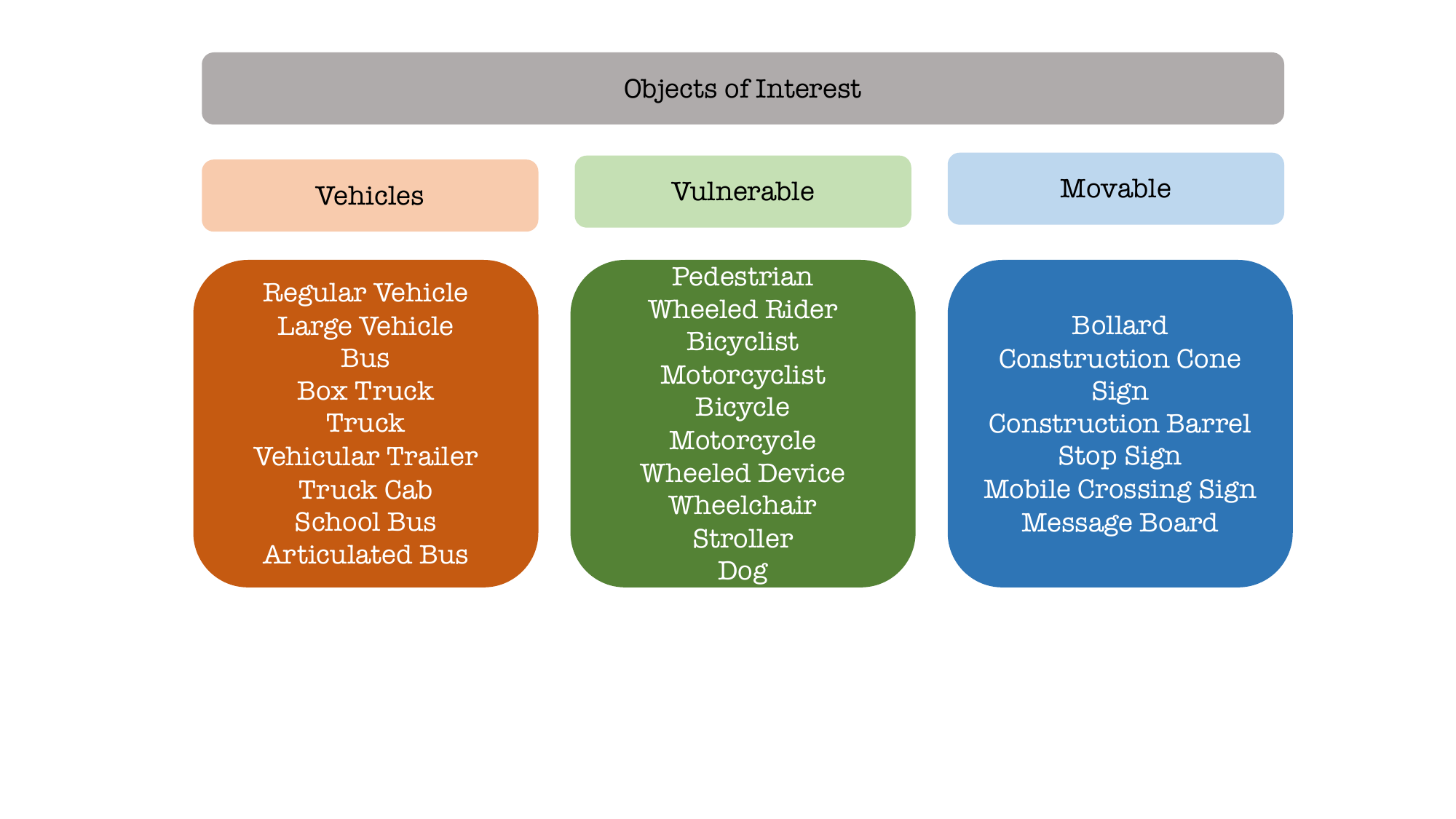} 
\caption{\small
{\bf Left:}
According to the histogram of per-class object counts (on the \textbf{left}), 
classes in Argoverse 2.0 (AV2) follow a long tailed distribution. 
Following~\cite{openlongtailrecognition} and nuScenes (Fig.~\ref{fig:histogram}), we report performance for three groups of classes based on their cardinality (split by dotted lines): {\tt Many}, {\tt Medium}, and {\tt Few}.
As AV2 does not provide a class hierarchy, we construct one  by referring to the nuScenes hierarchy (cf. Fig.~\ref{fig:av2_hierarchy} on the \textbf{right).}
}
\label{fig:av2_hierarchy}
\end{figure}

We present results on the large-scale Argoverse 2.0 (AV2) dataset, another long-tailed dataset developed for autonomous vehicle research (Fig.~\ref{fig:av2_hierarchy} on the \textbf{left}). 
AV2 evaluates on 26 classes, which follow the long-tailed distribution. As AV2 does not provide a semantic hierarchy, we construct one (cf. Fig.~\ref{fig:av2_hierarchy} on the \textbf{right}) by adapting the nuSccenes hierarchy.
As show in Table~\ref{tab:av2}, our main conclusions still hold on AV2.
Compared to the CenterPoint reported~\cite{wilson2021argoverse}, our implementation improves mAP by 14.4 by carefully adopting LiDAR sweep aggregation and class-balanced sampling. 
Based on our implementation, exploiting hierarchical semantics improves mAP from 44.0 to 47.0. This improves performance for both {\tt common} and {\tt rare} classes alike. 
Interestingly, FCOS3D performs significantly worse than the LiDAR based detectors, yet multimodal filtering improves mAP to 48.4 mAP. These new results on AV2 are consistent with those on nuScenes, demonstrating the general applicability of our approach.

{
\setlength{\tabcolsep}{0.7em} 
\begin{table}[h]
\small
\centering
\caption{\small 
Results (mAP in \%) on Argoverse 2.0 (AV2) evaluated at 50 meters. 
Compared to the CenterPoint reported in~\cite{welchman2004human}, our implementation improves mAP by 14.4 by carefully adopting LiDAR sweep aggregation and class-balanced sampling. 
Based on our implementation, exploiting hierarchical semantics improves mAP from 44.0 to 47.0. 
Note that AV2 does not provide a semantic hierarchy; we construct one based on the nuScenes hierarchy (Fig.~\ref{fig:hierarchy}).
Interestingly, FCOS3D performs significantly worse than the LiDAR based detectors, yet multimodal filtering improves mAP to 48.4 mAP. These new results on AV2 are consistent with those on nuScenes (cf. Table~\ref{tab:lt3d}), demonstrating the general applicability of our approach.
}
\begin{tabular}{l c c c c c c c c c c c}
\toprule
\multirow{1}{*}{Method} & \multirow{1}{*}{Multimodal} & \multicolumn{1}{c}{\tt Many} & \multicolumn{1}{c}{\texttt{Medium}} & \multicolumn{1}{c}{\texttt{Few}} & \multicolumn{1}{c}{\tt All} \\ 

\midrule
\multirow{1}{*}{FCOS3D \cite{wang21_fcos3d} (RGB-only)}                &           &     27.4       &    17.0       &     7.8      &      14.6        \\ 
\midrule
\multirow{1}{*}{CenterPoint (LiDAR-only) \cite{yin2020center} reported by \cite{wilson2021argoverse}}   &            &    66.7       &   32.9        &    14.1         &   29.6       \\
\multirow{1}{*}{CenterPoint (LiDAR-only) [Our Implementation]}         &           &    77.4        &       46.9    &   30.2        &     44.0            \\ 
\multirow{1}{*}{\quad\quad + Hierarchy}                  &            &    {\bf 79.0}       &   50.3        &      33.6       &   47.0       \\
\multirow{1}{*}{\quad\quad + Multimodal Filtering}    &      \checkmark      &   {\bf 79.0}        &       {\bf 51.4}    &       {\bf 35.3}      &     {\bf 48.4}     \\
\bottomrule
\end{tabular}
\label{tab:av2}
\end{table}
}



\end{document}


\maketitle
\newcommand{\problem}{LT3D}
\newcommand{\eg}{e.g.}

\noindent{\bf Outline: }
In this supplementary document, we first study class confusion within coarse grained categories. Then, we analyze the challenge of applying off-the-shelf multimodal methods such as MVP to \problem. 
Lastly, we explore another heuristic method for multimodal fusion that  incorporates detection scores, validating that multimodal cues are necessary for \problem.

\section{Analysis of Misclassifications with Confusion Matrix}

\begin{figure*}[h]
\centering
\hspace{-5mm} {\tt Vehicle} \hspace{27mm} {\tt Pedestrian} \hspace{27mm} {\tt Movable}\\
\includegraphics[width=0.32\linewidth,valign=t]{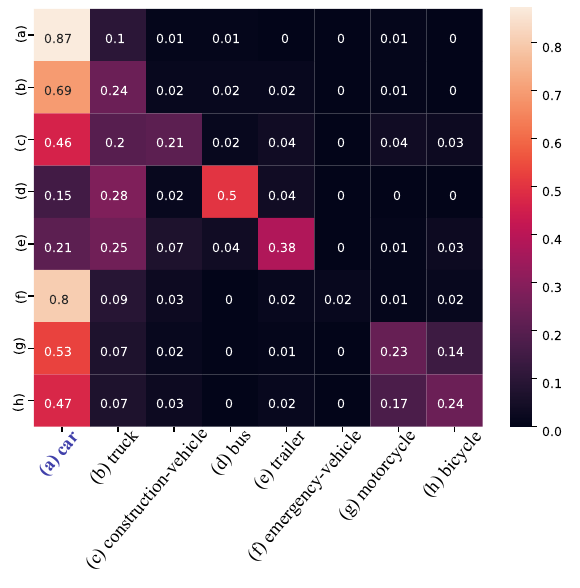} \hfill
\includegraphics[width=0.32\linewidth,valign=t]{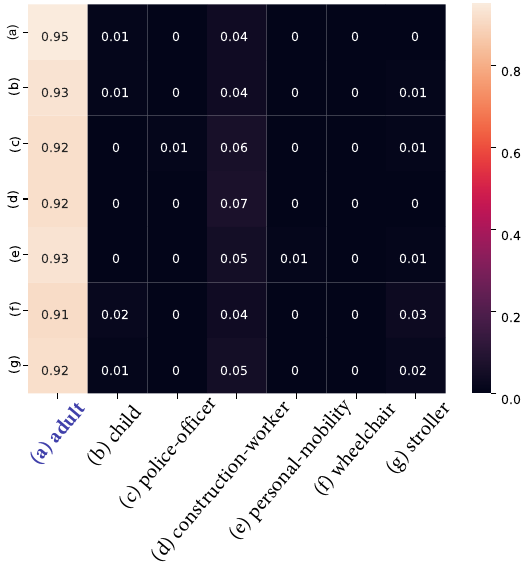} \hfill
\includegraphics[width=0.32\linewidth,valign=t]{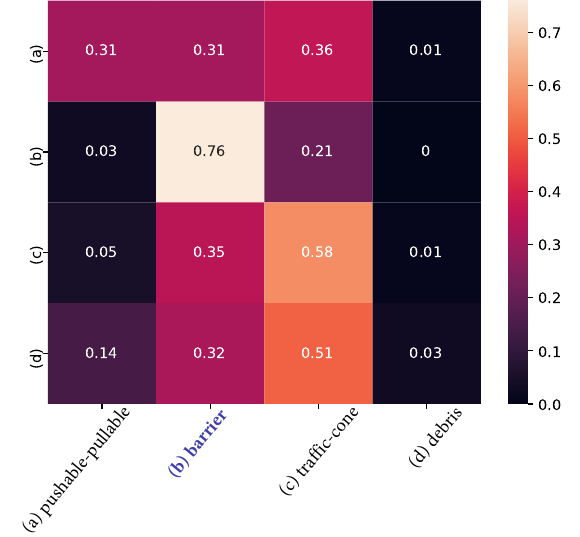}\hfill
\vspace{-3mm}
\caption{\small 
{\bf Breakdown analysis of misclassifications within superclasses.} We analyze our best-performing model (CenterPoint w/ hierarchical training and multimodal filtering). 
Fine-grained classes are most often confused by the dominant class (in {\bf \color{blue}blue}) in each superclass: ({\bf left}) {\tt Vehicle} is dominated by {\tt car}, ({\bf mid}) {\tt Pedestrian} is dominated by {\tt adult}, and ({\bf right}) {\tt Movable} is dominated by {\tt barrier}.
We find that class confusions are reasonable. {\tt Car} is often mistaken for {\tt truck}. Similarly, {\tt truck}, {\tt construction-vehicle} and {\tt emergency-vehicle} are most often mistaken for {\tt car}. 
{\tt Bicycle} and {\tt motorcycle} are sometime misclassified as {\tt car}, presumably because they are sometimes spatially close (within the 2m match threshold) to {\tt cars}.
{\tt Adults} have similar appearance to {\tt police-officer} and {\tt construction-worker}, and they are often co-localized with {\tt child}, {\tt wheelchair} and {\tt stroller}; all these might cause significant class confusion.
}
\label{fig:conf_matrix}
\end{figure*}

For 3D detection, localization and classification are two important measures for 3D detection performance. In practice, we cannot achieve perfect performance for either. 
In safety-critical applications, detecting but misclassifying objects (as a semantically related category) is more desirable than a missed detection (e.g., detect but misclassify a {\tt child} as {\tt adult} versus not detecting this {\tt child}). 
Therefore, we introduce hierarchical AP (AP$_H$), which  considers such semantic relationships across classes to award partial credit (details in the main paper). Applying this hierarchical AP reveals that classes are most often misclassified as their LCA=1 siblings within coarse-grained superclasses.
In this document, we use confusion matrices to further analyze the misclassifications within superclasses, as shown in Fig.~\ref{fig:conf_matrix}. 
Below we explain how to compute a confusion matrix for the detection task.

For each superclass, we make a confusion matrix, in which the entry $(i,j)$ indicates the misclassification rate of class-$i$ objects as class-$j$.
Specifically, given a fine-grained class $i$, we find its predictions that match ground-truth boxes within 2m center-distance of class-$i$ and all its sibling classes (LCA=1, within the corresponding superclass); we ignore all unmatched detections.
This allows us to count the mis-classifications of class-$i$ objects into class-$j$, with which a simple normalization produces misclassification rates. 


\section{Why Do State-of-the-Art Multimodal Detectors Perform Poorly for \problem?}

{
\setlength{\tabcolsep}{0.50em} 
\begin{table}[t] 
\small
\centering
\caption{\small 
Results (AP in \% on nuImages) of CenterNet2~\cite{zhou2021probablistic} (a crucial component in MVP~\cite{yin2021multimodal}) when adapted for \problem.
The adapted CenterNet2 (``for \problem'') performs consistently worse than its ``original'' version because it has to strike a balance between more classes during training.
Recall that since MVP uses CenterNet2's instance segmentation masks to densify LiDAR scans, false-positive masks will degrade detection performance. CenterNet2 produces near zero performance on rare classes such as {\tt emergency-vehicle} ({\tt EV}), {\tt wheelchair}, and {\tt personal-mobility} ({\tt PM}).
Note, {\tt TC} is {\tt traffic-cone}, {\tt CV} is {\tt construction vehicle}, {\tt MC} is {\tt motorcycle}, {\tt PP} is {\tt pushable-pullable}, {\tt CW} is {\tt construction-worker}, and {\tt PO} is {\tt police-officer}. We highlight classes with {\tt Medium} and {\tt Few} examples per class in {\color{blue}blue}. Even for 2D segmentation, we find that vulnerable classes like {\tt child} have low AP, which are aggregated into the {\tt pedestrian} superclass in the standard setup shown in {\color{darkgreen}green}. 
}
\begin{tabular}{l c c c c c c c c c c c c c}
\toprule
& \multicolumn{1}{c}{Car} & \multicolumn{1}{c}{Barrier} & \multicolumn{1}{c}{TC} & \multicolumn{1}{c}{Truck} &  \multicolumn{1}{c}{\color{blue}Trailer} & \multicolumn{1}{c}{\color{blue}Bus} & \multicolumn{1}{c}{\color{blue}CV} &  \multicolumn{1}{c}{\color{blue}MC} & \multicolumn{1}{c}{\color{blue}Bicycle} & \multicolumn{1}{c}{\color{darkgreen}Pedestrian}  \\ 

\midrule
Original &  58.6   &  52.0   &  47.9   &  51.5   &  21.4   & 53.4   &  27.6   &  49.9   &  33.6  & 37.8     \\ 
\midrule
\multirow{1}{*}{for LT3D}                     &  57.4   &  50.2  &  46.6   &  50.0   &  18.8   &  51.2   &  25.1   &  48.2   &  33.0  & ---       \\ 
\midrule
\midrule 
\multirow{1}{*}{} & \multicolumn{1}{c}{Adult} & \multicolumn{1}{c}{\color{blue}PP} & \multicolumn{1}{c}{\color{blue}CW} & \multicolumn{1}{c}{\color{blue}Debris} &  \multicolumn{1}{c}{\color{blue}Child} & \multicolumn{1}{c}{\color{blue}Stroller} & \multicolumn{1}{c}{\color{blue}PO} &  \multicolumn{1}{c}{\color{blue}EV} & \multicolumn{1}{c}{\color{blue}Wheelchair} & \multicolumn{1}{c}{\color{blue}PM} \\ 

\midrule
Original        & --- & --- & --- & --- & --- & --- & --- & --- & --- & ---        \\
\multirow{1}{*}{for LT3D}                     &  35.8  &  23.8 &  26.9   &  5.0  &  4.3   &  19.6   &  10.2   &  0.4   &  0.7    &  1.01       \\ 
\bottomrule
\end{tabular}
\label{tab:segm}
\vspace{-1mm}
\end{table}
}

Multimodal  3D detectors can leverage rich image features from RGB and metric depth provided by lidar, facilitating better detection performance compared to prior single-modal methods on standard benchmarks. 
We test two state-of-the-art multimodal 3D detectors (TransFusion~\cite{bai22transfusion} and MVP~\cite{yin2021multimodal}) by adapting them for \problem.
However, they did not perform as well as expected. Surprisingly, the lidar-only CenterPoint detector performed the best. We analyze this surprising result below.

TransFusion uses a DETR-like architecture to train an end-to-end transformer-based detector. However, like DETR, the number of predictions TransFusion can make per sweep is limited by the number of object queries (set at 200). As the number of objects in a sweep approaches the number of queries, performance saturates and the detector misses more objects \cite{detr2020carion}. 
Moreover, transformers are difficult to train in practice. Though we tried our best to tune hyperparameters when adapting TransFusion for \problem, we find it non-trivial. We expect that more effort is required to improve TransFusion for \problem.

MVP extends CenterPoint by using an off-the-shelf instance segmentation model to densify regions of point clouds with image evidence. However, retraining instance segmentation models for \problem\ results in low performance for {\tt rare} classes. Importantly, this low segmentaiton performance results in many false positive masks, adding noise to the densified LiDAR. As shown in Table \ref{tab:segm}, {\tt common} class performance for CenterNet2 trained for \problem \ consistently performs worse than the OTS model. Moreover, many {\tt rare} classes  including {\tt debris}, {\tt child}, {\tt stroller}, {\tt emergency-vehicle} and {\tt personal-mobility} have near zero IOU. Importantly, these false positive detections from {\tt rare} classes may negatively impact {\tt common} classes as these {\tt rare} classes may provide incorrect semantic information.

\section{Multimodal Fusion via Rescoring} 

Detecting {\tt rare} classes from LiDAR alone (i.e., sparse point clouds) is challenging, whereas exploiting multimodal cues, particularly high-resolution RGB images, can improve 3D detection of {\tt rare} classes, as demonstrated by our multimodal filtering method in the main paper. We explore an additional simple fusion method to further support the utility of multimodal fusion for \problem. Briefly, this method rescores LiDAR-based detections by exploiting RGB-based detections, detailed below.

Our rescoring method exploits RGB-based 2D detections to rescore LiDAR-based 3D detections.
Intuitively, if both RGB- and LiDAR-based detections fire on the same object, we should upweight the confidence score for the LiDAR-based 3D detection;
if there is no image evidence for an object but only the LiDAR detector fires, we should downweight the confidence of the LiDAR-based detection. 
In this work, we test the rescoring technique with our best performing 3D detectors (cf. the LiDAR-only CenterPoint detector \cite{yin2020center} and multimodal TransFusion detector \cite{bai22transfusion}), and the RGB-based 2D detector CenterNet2 \cite{zhou2021probablistic}.

\begin{figure}[t]
\centering
\small
\hspace{5mm} LiDAR-based Detections \hspace{17mm} RGB-based Detections \hspace{25mm} Rescoring \hspace{15mm} \ \\
\includegraphics[width=0.32\linewidth]{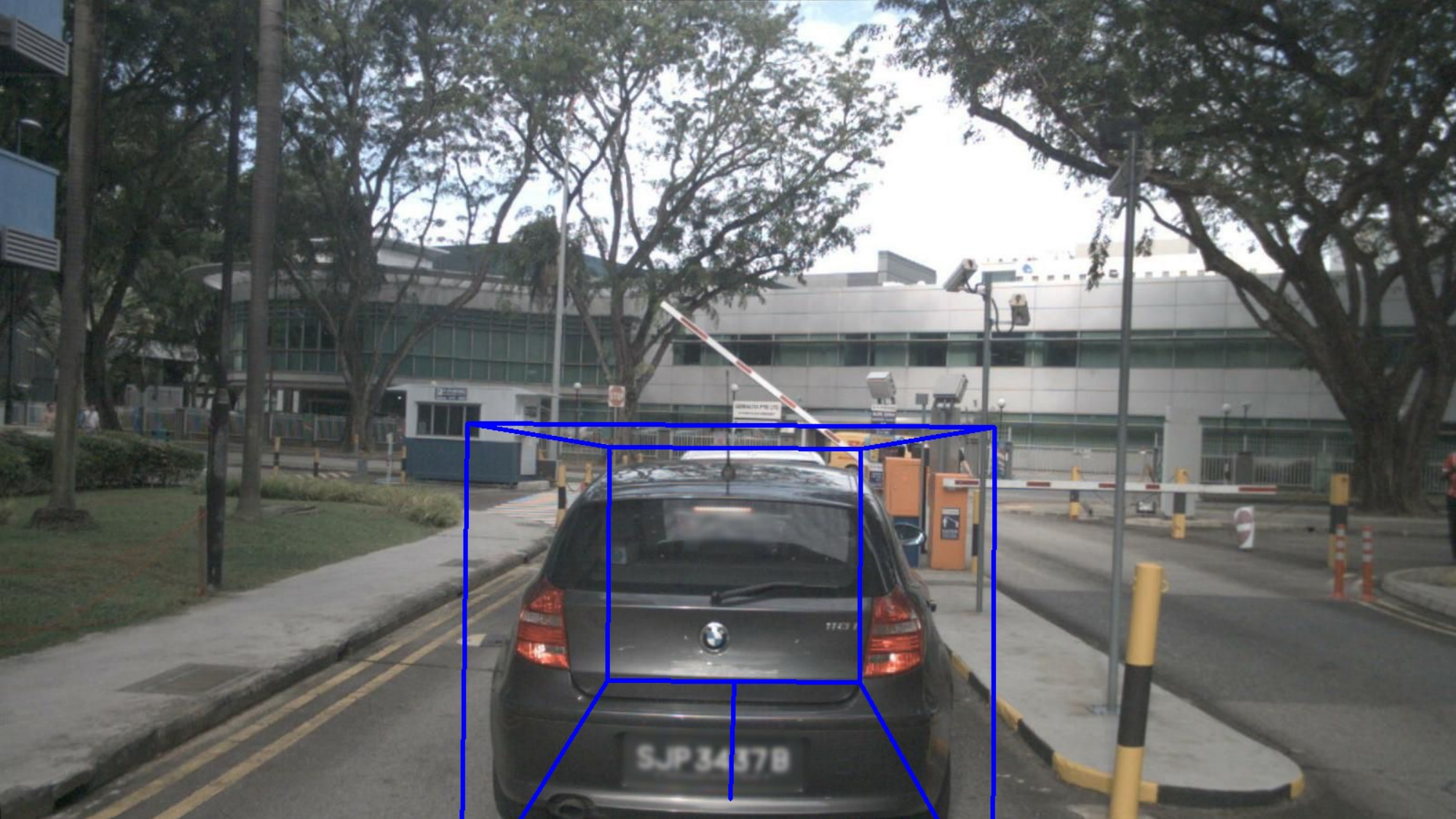} 
\includegraphics[width=0.32\linewidth]{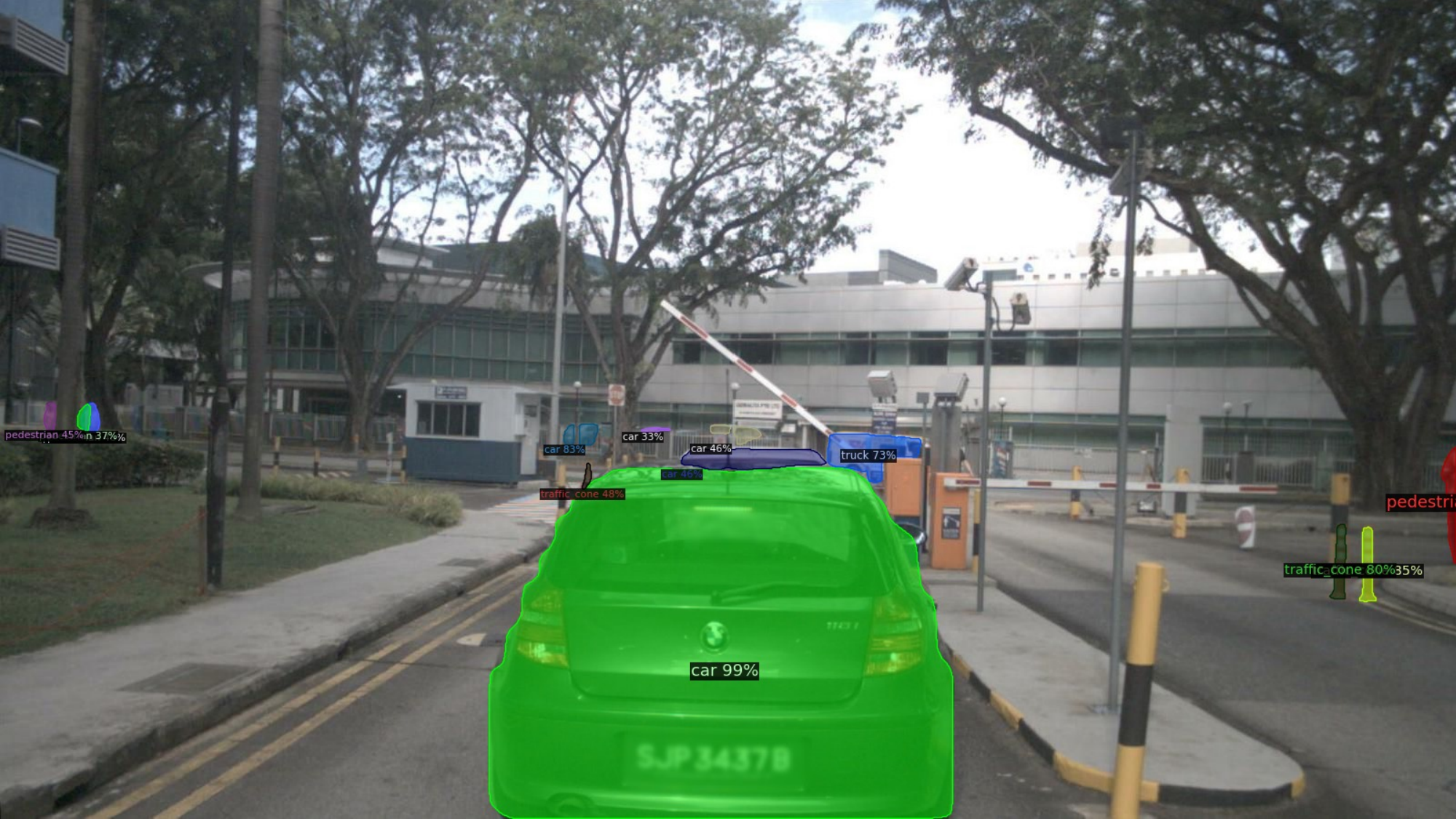} 
\includegraphics[width=0.32\linewidth]{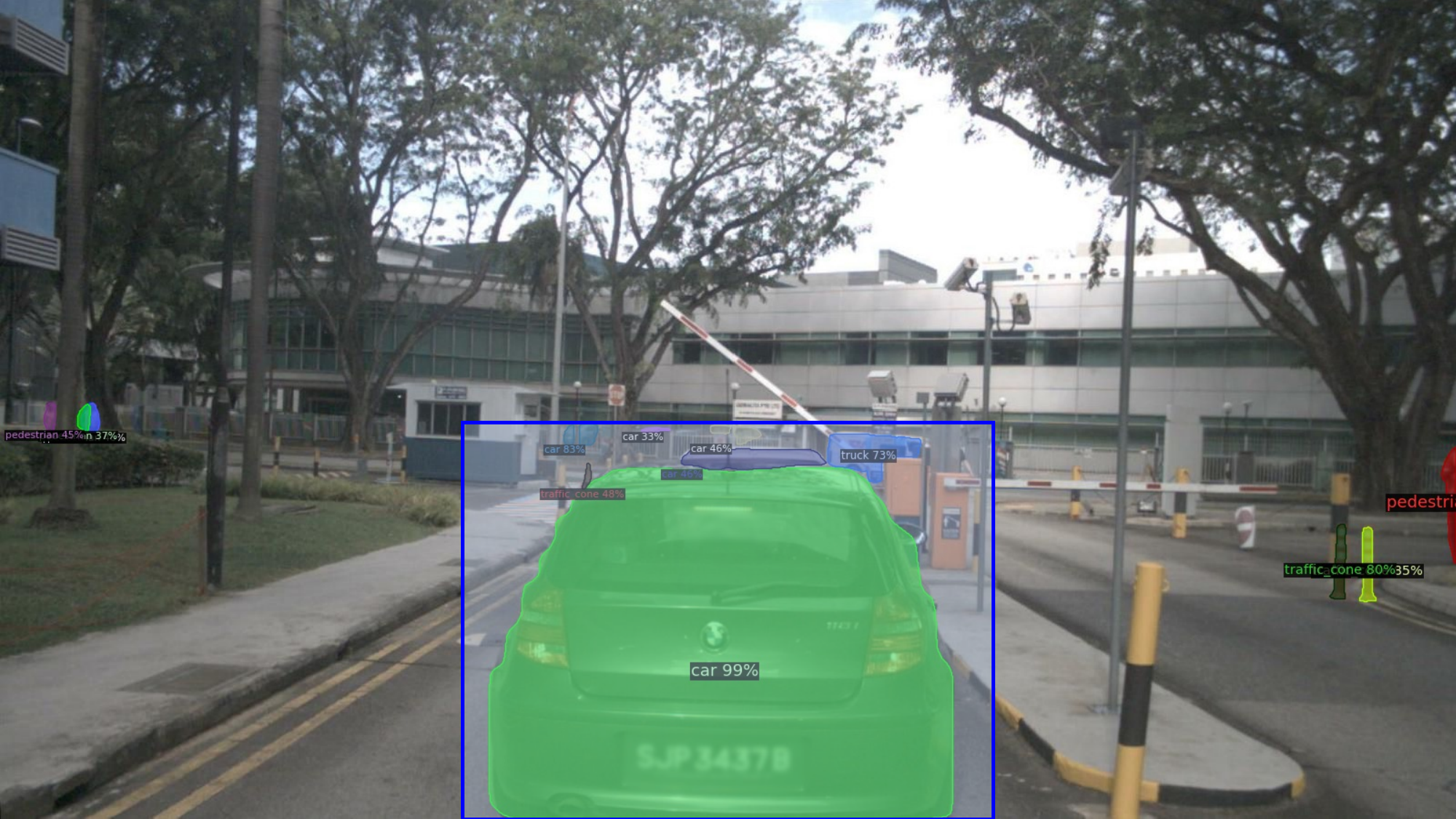}
\caption{\small
Illustration of the proposed rescoring fusion technique, which upweights LiDAR-based detections based on image evidence. 
Given some 3D detections from a LiDAR detector, we project them onto the 2D image plane and consider how the projected 2D boxes overlap any 2D detections. Intuitively, if they overlap, we should upweight the confidence score of the LiDAR-based detection by $1.25$x. Similarly, if there is no image evidence for an object, we should downweight score by $0.75$x.
} 
\label{fig:filtering}
\end{figure}

Figure~\ref{fig:filtering} illustrates the implementation of rescoring. We project 3D detections (e.g., from CenterPoint) onto the 2D image plane, and consider the overlapping region between the projected 2D bounding box and any of the 2D detections (e.g., from CenterNet2~\cite{zhou2021probablistic}). If they match, we claim that they fire on the same object and we heuristically increase the confidence score by $1.25$x.
For LiDAR-based detections that do not match any 2D RGB detections, we do not modify their scores.\shu{the caption  of Fig 2 says 0.75x!}

We show results of rescoring, along with filtering, in Table~\ref{tab:rescoring}.
Both of our multimodal fusion methods improve SOTA detectors for \problem, and achieve similar performance gain.
Using the fusion methods together (``w/ Filtering + Rescoring'') yields marginal improvement for CenterPoint but not for TransFusion. Importantly, we achieve the best performance by applying our multi-modal fusion method to the lidar-based CenterPoint detector.

{
\setlength{\tabcolsep}{1.0em} 
\begin{table}[b]
\small
\centering
\caption{\small 
We test our two post-hoc multimodal fusion methods when applied to state-of-the-art (SOTA) detectors, namely the lidar-based detector CenterPoint and the multimodal detector TransFusion.
Results (mAP in \% on nuScenes) are comparable to Table 1 in the main paper.
Both our fusion methods improve SOTA detectors for \problem, and achieve similar performance gain.
Using the fusion methods altogether (``w/ Filtering + Rescoring'') yields marginal improvement for CenterPoint but not for TransFusion. 
}
\begin{tabular}{l c c c c c c c c c c c}
\toprule
\multirow{1}{*}{Method} & \multirow{1}{*}{Multimodal} & \multicolumn{1}{c}{\tt Many} & \multicolumn{1}{c}{\texttt{Medium}} & \multicolumn{1}{c}{\texttt{Few}} & \multicolumn{1}{c}{\tt All} \\ 

\midrule
\multirow{1}{*}{CenterPoint (LiDAR-only)}                       &            &  73.7             & 41.3         &   3.0        &   37.5        \\ 
\multirow{1}{*}{CenterPoint (LiDAR-only) + Hierarchy}          &                                    &  \textbf{77.1}             & 45.1         &   4.3         &   40.4      \\
\multirow{1}{*}{\quad \quad $\indent$ w/ Filtering}         &   \checkmark                       &  \textbf{77.1}             & 49.0          &   9.4      &   43.6       \\ 
\multirow{1}{*}{\quad\quad  $\indent$ w/ Rescoring}                                &  \checkmark      &  76.8           &    \textbf{49.8}      &     5.7      &   42.6   \\ 
\multirow{1}{*}{\quad\quad  $\indent$ w/ Filtering + Rescoring}                    &  \checkmark      &  76.7           &    49.0      &     \textbf{10.0}      &   \textbf{43.7}   \\

\midrule
\multirow{1}{*}{TransFusion (LiDAR-only)}                   &                  &  68.5            & \textbf{42.8}          &   8.4        &   38.5       \\ 
\multirow{1}{*}{TransFusion (LiDAR+RGB)}                                    &  \checkmark      &  \textbf{73.9}            & 41.2          &   \textbf{9.8}        &   39.8       \\
\multirow{1}{*}{\quad \quad $\indent$ w/ Filtering}                                &  \checkmark      &  \textbf{73.9}            & 42.5          &   9.1        &  40.1      \\ 
\multirow{1}{*}{\quad\quad  $\indent$ w/ Rescoring}                                &  \checkmark      &    73.5         &    42.7      &   \textbf{9.8}        &  \textbf{40.3}    \\ 
\multirow{1}{*}{\quad\quad  $\indent$ w/ Filtering + Rescoring}                    &  \checkmark      &     73.5        &   42.6      &    8.9        &   40.0   \\ 

\bottomrule
\end{tabular}
\label{tab:rescoring}
\end{table}
}

\section{Open-Source Code for Methods and Evaluation}
Our extensive study shows that this new problem is very challenging and requires further exploration in the community. We will release open-source code to foster this exploration.
As a part of supplementary material, we include both the code for our baseline methods and evaluation protocol. Our methods build on top of the CenterPoint codebase, and our evaluation code extends the official nuScenes devkit. 

\bibliography{references}  